\documentclass[sigconf,nonacm]{acmart}
\makeatletter
\@ACM@balancefalse
\makeatother

\usepackage{tikz}
\usetikzlibrary{shapes.geometric, arrows.meta, positioning, fit, backgrounds, calc, chains}
\usepackage{xcolor}
\usepackage{placeins}
\usepackage{array}

\newcommand{\sysname}{ACES}
\newcommand{\sysnamefull}{Agentic Continuous Evaluation of Skills}
\newcommand{\expectedbehavior}{\texttt{expected\_}\allowbreak\texttt{behavior}}
\newcommand{\authorsep}{,\ }

\tikzset{
  layer/.style       = {rectangle, rounded corners=2pt, draw,
                         minimum width=2.4cm, minimum height=0.8cm,
                         align=center, font=\small},
  evaluator/.style   = {rectangle, rounded corners=2pt, draw,
                         minimum width=2.2cm, minimum height=0.7cm,
                         align=center, font=\footnotesize},
  dimension/.style   = {rectangle, rounded corners=2pt, draw,
                         minimum width=1.8cm, minimum height=0.6cm,
                         align=center, font=\footnotesize, dashed},
  artifact/.style    = {rectangle, draw,
                         minimum width=2.0cm, minimum height=0.7cm,
                         align=center, font=\footnotesize},
  bigbox/.style      = {rectangle, rounded corners=3pt, draw, thick,
                         inner sep=4pt, align=center},
  flow/.style        = {-Latex, thick},
  flowlight/.style   = {-Latex, thin, gray},
}

\setcopyright{none}
\begin{document}

\title{Evaluating Skills, Not Just Agents: Agentic Continuous Evaluation of Skills}

\author{%
  {\large Christopher Kevin\textsuperscript{*}\authorsep Narendran Raghavan\authorsep
  Jean-Francois Puget\authorsep Roshni Malani\authorsep Meghana Puvvadi\\
  Moshe Abramovitch\authorsep Mohit Gupta\authorsep Rama Akkiraju\authorsep
  Subodh Prabhu\authorsep Yogesh Dangi\authorsep Wei Luo\authorsep Seong Hee Lee}\\[0.45em]
  {\normalsize\normalfont NVIDIA}\\[-0.05em]
  {\normalsize\normalfont\href{mailto:christopherk@nvidia.com}{christopherk@nvidia.com}}%
}

\makeatletter
\gdef\authors{Christopher Kevin, Narendran Raghavan, Jean-Francois Puget,
  Roshni Malani, Meghana Puvvadi, Moshe Abramovitch, Mohit Gupta,
  Rama Akkiraju, Subodh Prabhu, Yogesh Dangi, Wei Luo, Seong Hee Lee}
\makeatother

\renewcommand{\shortauthors}{Kevin et al.}

\begin{abstract}
Enterprise agent programs are moving from prototypes into production,
where reusable skills, tools, and workflow packages must be reviewed
with evidence rather than prose. Current gates often scan these
artifacts for structure, style, and security, but they do not answer
the deployment question: does the capability package help a live agent
complete enterprise tasks under the same model, sandbox, and grading
policy?

We present \sysname{} (\sysnamefull{}), a repository-native framework
for evaluating skills and product capability packages as executable
agent artifacts. \sysname{} runs paired live trials with and without a
target skill, normalizes trajectories into the Agent Trajectory
Interchange Format (ATIF), grades six default runtime metrics, and
reports \emph{Skill Lift}: the target skill's added value for a fixed
task, harness, workspace, and scorer. The same protocol supports
product-owned task suites that compare baseline, skill, bundle,
team-skill, and plugin targets.

On 145 real skills from internal enterprise repositories and public
catalogs, scan-only gates surface useful authoring issues but measure
complementary facets (structural versus LLM-judge Spearman
$\rho=0.14$). Across 947 scored paired cases from 58 of 64 production
skills and four primary harnesses, mean composite Skill Lift is 0.2134
(95\% paired-case CI [0.1967, 0.2301]); mean outcome-only lift, the
average of accuracy and goal accuracy, is 0.1799. Composite lift is
positive in 72.8\% of paired cases. The largest process-metric gains
appear in skill execution, behavior check, and skill efficiency---signals
about discovery, routing, workflow following, and tool use that document
scans cannot observe.
An open-source implementation of the methodology is available in NVIDIA
SkillEvaluator.
\end{abstract}



\maketitle

\section{Introduction}
\label{sec:intro}

\noindent\textbf{Skills are the application layer for agents.}
Over the past year, LLM-based agents have acquired a repeatable
extension mechanism: \emph{agent skills}. A skill is a
natural-language package of procedural know-how---a
\texttt{SKILL.md} description, optional deterministic scripts the
agent invokes for steps it should not improvise, plus references and
examples---that the agent loads at inference time. The defining design
choice is progressive disclosure: the agent reads only the short
description up front, and pulls in instructions, scripts, or examples
only when it decides the skill applies, so dozens of skills can
coexist in a workspace without bloating context~\cite{anthropic-skills-engineering,
anthropic-skills-bestpractices, openai-eval-skills,
vercel-skills-faq}. The format is adopted across Claude Code, Codex,
Cursor, and other agent harnesses, and
community registries already host thousands of user-contributed
skills. Bundled sets of related skills are increasingly called
plugins; the methodology in this paper applies to a single skill, a
plugin, or any composition of skills an agent can load.

\smallskip
\noindent\textbf{The current state of practice.}
Industry tooling for skill evaluation today clusters into four
complementary classes.
(1)~\emph{Structural checks} enforce a skill specification---frontmatter
fields, section layout, script presence, naming
conventions---and produce a deterministic score~\cite{skillsbench,
anthropic-skillcreator}.
(2)~\emph{LLM-as-Judge rubrics} grade subjective qualities of the
documentation that static rules cannot see: clarity of instructions,
scope definition, example quality, trigger phrasing~\cite{geval,
confident-geval, langchain-evalskills}.
(3)~\emph{Linters} check the scripts themselves for syntax, style,
and dangerous patterns.
(4)~\emph{Security scanners} detect prompt-injection markers,
leaked secrets, destructive shell patterns, and suspicious URL
references inside the skill content.
All four classes share a structural property: they \emph{scan the
skill document}. None runs it.

\smallskip
\noindent\textbf{Scanning is necessary, not sufficient.}
Running a skill through scan-only evaluation is analogous to
compiling a program with \texttt{-Wall -Werror}: the absence of
warnings does not mean the program does what it should. A skill
can scan cleanly and still fail at runtime because
(a)~the agent never discovers it when the user asks a relevant
question,
(b)~the agent reads the documentation but invokes the wrong script
or wrong arguments,
(c)~the agent produces correct output but misinterprets and
misreports it,
(d)~the skill collides with another skill already in the workspace,
or
(e)~the skill is silently regressed by a model update that changes
how the agent reasons about its documentation.
None of these failure modes is observable from the skill alone.

\smallskip
\noindent\textbf{Measuring the gap.}
We evaluated 145 real skills from internal enterprise repositories
and public catalogs with both structural and LLM-judge tiers. 94.5\% of
skills pass the default C-grade gate on structural checks, and
86.2\% pass the LLM-judge rubric. And yet the two scores correlate
at Spearman $\rho = 0.14$. In other words, even among the
scan methods \emph{the two disagree with each other}. This is a
sharp quantitative cue that doc-scanning is not converging on a
single coherent notion of skill quality---and neither method tells
us what the agent actually does with the skill at runtime.

\smallskip
\noindent\textbf{\sysname{}: extending static skill-scanning with
live agent evaluation.}
We present \sysname{} (\sysnamefull{}), a methodology and system for
evaluating skills as executable agent artifacts rather than static
documents alone. \sysname{} is organized around three portable
interfaces: an evaluation-asset contract authored with the skill, an
adapter that materializes those assets into paired runtime tasks, and
an ATIF-based trajectory contract that lets the same grading layer
compare behavior across agent harnesses. For each author-provided
task, \sysname{} runs paired conditions: a with-skill condition, where
the target skill is available, and a baseline condition, where the
target skill is withheld while configured prerequisite, helper,
reference, or decoy skills remain fixed. The paired difference yields
\emph{Skill Lift}: the marginal value contributed by the target skill
under a fixed agent, model, task, workspace, and grading policy.
The \sysname{} live-evaluation methodology is available in NVIDIA
SkillEvaluator, an open-source multi-tier framework whose Tier~3
operationalizes the paired evaluation protocol described in this
paper~\cite{skillevaluator}.
Our contributions:
\begin{enumerate}
  \item A methodology for skill-as-artifact evaluation that
    extends doc-scanning with live agent evaluation on author-owned
    tasks, including paired with-skill/baseline measurement and Skill
    Lift (\S\ref{sec:scanning}, \S\ref{sec:live-eval}).
  \item An evaluation-asset authoring workflow in which authors can
    write datasets directly, seed or refine cases with LLM assistance
    and real trajectories, use free-form
    \expectedbehavior{} assertions for LLM-judged workflow
    checks, and attach BYOT/BYOG assets when generic metrics are
    insufficient (\S\ref{sec:dataset}, \S\ref{sec:refine}).
  \item A shared ATIF-based grading layer that applies deterministic,
    security, LLM-judge, and optional domain-specific metrics to
    trajectories across agent harnesses, with stakeholder-facing
    dimensions for author review
    (\S\ref{sec:evaluators}--\S\ref{sec:atif}).
  \item A dynamic \sysname{} adapter that stages fresh paired task
    environments across agents, task sources, workspace modes,
    grading modes, and sandbox backends, including isolated and group
    workspaces for skill-selection and prerequisite-skill scenarios
    (\S\ref{sec:harbor}--\S\ref{sec:isolation}).
  \item An evaluation-native skill-development workflow in which
    structural checks, LLM-judge scoring, security scanning, optional
    live-agent evaluation, and Skill Lift reports become review
    evidence for skill changes across product repositories,
    skill-centric repositories, and registries
    (\S\ref{sec:discipline}).
	  \item An empirical study on 145 real skills showing that scan-only
	    signals are incomplete: 94.5\% pass the default structural gate,
	    and deterministic and LLM-judge scores correlate weakly at
	    Spearman $\rho = 0.14$. Across 947 scored paired cases from 58 of
	    64 production skills and four primary harnesses, mean composite
	    Skill Lift is 0.2134 (95\% paired-case CI [0.1967, 0.2301]) and
	    mean outcome-only lift is 0.1799. The same-agent model slice shows
	    why paired measurement matters: absolute scores can rise while
	    marginal skill lift shrinks as the baseline model improves
	    (\S\ref{sec:empirical}).
	\end{enumerate}

We defer related work to \S\ref{sec:related} so the methodology
and measured results land first.

\section{Evaluating the Skill Artifact}
\label{sec:scanning}

This section covers the four scan-only evaluation classes that
exist today, describes what each adds beyond the others, and
measures where they collectively leave an observability gap on
our 145-skill corpus (\S\ref{sec:scanlimits}).

\subsection{Static Structural Quality}
\label{sec:static}

Deterministic checks against a skill specification are the
cheapest and most reproducible first pass: no API key, runs in
milliseconds, gating-friendly. Our implementation runs roughly
fifty rule checks across four weighted dimensions that each
start at 100 and deduct on every finding. The check families are
summarized in Table~\ref{tab:static-rules}; each rule in a family
independently contributes a severity-weighted deduction. The
overall score is a weighted sum of the four dimensions; the
default gate accepts scores at 70 or above. The rules are closely
aligned with vendor guidance on how skills should be
written~\cite{anthropic-skills-bestpractices,
anthropic-skillcreator, claude-skills-howto}.

\begin{table}[t]
\small
\centering
\caption{Static check families per dimension
(${\approx}50$ rules total). Each family contains multiple rules
that fire independently; each rule deducts from its dimension
score.}
\label{tab:static-rules}
\begin{tabular}{@{}p{0.32\linewidth}p{0.62\linewidth}@{}}
\toprule
\textbf{Dimension (weight)} & \textbf{Rule families} \\
\midrule
Correctness (0.35)     & Frontmatter contract (\texttt{name},
                         \texttt{description}, \texttt{version},
                         \texttt{author}, \texttt{tags},
                         \texttt{tools}); name/directory match;
                         \texttt{Instructions} section;
                         \texttt{run\_script} mention;
                         script-file presence; header-table
                         documentation; injection-suspicious
                         characters. \\
\addlinespace[2pt]
Discoverability (0.25) & Description length (50--150 preferred; $>$200 flagged as overlong);
                         trigger vocabulary
                         (\emph{use}, \emph{when}, \emph{for});
                         avoidance of vague wording;
                         explicit negative/boundary phrasing;
                         directory-name conventions; tone
                         (first/second person); ``WHEN to use''
                         guidance. \\
\addlinespace[2pt]
Reliability (0.25)     & Error-handling vocabulary;
                         input-validation hints in
                         \texttt{scripts/*}; \texttt{Prerequisites},
                         \texttt{Limitations},
                         \texttt{Troubleshooting} sections;
                         MCP-connection guidance when MCP is
                         declared. \\
\addlinespace[2pt]
Efficiency (0.15)      & Token budget; repeated lines; list-format
                         instructions; nested-reference depth;
                         corporate/hedge language;
                         oversize-skill warnings. \\
\bottomrule
\end{tabular}
\end{table}

\subsection{LLM-as-Judge Clarity: What Static Cannot See}
\label{sec:rubric}

Structural rules cannot judge whether an instruction is
\emph{clearly phrased}, whether an example is
\emph{representative}, or whether a description would actually
\emph{trigger} the skill on a realistic user query. We therefore
add a second, subjective layer grounded in the G-Eval form-filling
approach~\cite{geval, confident-geval}. A judge model reads
\texttt{SKILL.md}, optional script content, and the structural
results (as context), then returns a JSON object with per-criterion
pass/fail and 0--10 scores. Temperature is fixed at zero. We use
nine base criteria, plus a tenth (\emph{structural coherence},
implemented as \texttt{tier1\_coherence} in the code) that
activates when auxiliary script or reference content is supplied
(Table~\ref{tab:rubric}).

\begin{table}[t]
\small
\centering
\caption{LLM-as-Judge rubric criteria
(9 base + 1 conditional). Each is scored 0--10 with free-form
reasoning.}
\label{tab:rubric}
\begin{tabular}{@{}p{0.34\linewidth}p{0.60\linewidth}@{}}
\toprule
\textbf{Criterion} & \textbf{What the judge is asked} \\
\midrule
description clarity       & Does the description communicate what
                            the skill does in one pass, without
                            ambiguity? \\
instruction clarity       & Are the instructions specific, ordered,
                            and executable by an agent reading them
                            cold? \\
example quality            & Do worked examples cover the prompt
                            variations an agent will actually see? \\
documentation completeness & Are the scripts, inputs, and expected
                            outcomes documented? \\
scope definition          & Is the boundary between what this skill
                            does and what it does not explicit? \\
professional tone         & Is the writing direct, concise, and
                            free of marketing or hedge language? \\
trigger simulation        & Would a realistic user query actually
                            route to this skill given its
                            description? \\
workflow completeness     & Are the read-before-execute sequence,
                            error paths, and cleanup all present? \\
error-handling quality    & Are specific failure modes and the
                            agent's response to each described? \\
\addlinespace[2pt]
\textit{structural coherence} & Does the structural feedback
                            align with what the judge observes?
                            (Activated only when auxiliary content
                            is supplied.) \\
\bottomrule
\end{tabular}
\end{table}

The LLM-judge catches errors structural checks miss. An
instruction such as ``handle errors appropriately'' passes the
structural rule (the vocabulary is present) but fails
error-handling quality (no specific behavior described). A
description that is the right length still fails trigger
simulation if a realistic user query would not match. Conversely,
the judge can score a skill with incomplete frontmatter highly if
its content is excellent---which matters for the pivot
in~\S\ref{sec:scanlimits}.

\smallskip
\noindent\textbf{Cross-judge-model sensitivity.}
Absolute LLM-judge values depend on the judge model. We rotated
three judges---Claude 3.7~Sonnet, Claude 3.5~Haiku, and a 9B
in-house judge---across our 145-skill corpus. The spread between
the strictest and most lenient judge is roughly 1.5 points on
the 0--10 scale; relative ranking within a corpus is largely
preserved, but we flag judge attribution on any absolute
claim~\cite{geval}.

\subsection{Linting and Security Scanning}
\label{sec:lint}

Two additional scan classes read the scripts and the full
content rather than just \texttt{SKILL.md}. \emph{Script linting}
applies a Python linter to \texttt{scripts/*.py}, advisory but
valuable for author feedback. \emph{Security scanning} detects
prompt-injection markers (instructions that subvert the system
prompt), leaked secrets (API keys and tokens matching common
patterns), destructive shell invocations
(\texttt{rm\,-rf}, \texttt{DROP\,TABLE}, \texttt{curl\,|\,sh}),
suspicious URL references, and supply-chain red flags. A skill
may have a perfect structure and rubric score yet still contain a
prompt-injection vector in a reference file. Security scanning is
an external integration in our pipeline rather than a contribution
of this paper.

\subsection{What Doc-Scanning Tells Us---and What It Misses}
\label{sec:scanlimits}

\begin{figure*}[t]
\centering
\includegraphics[width=\textwidth]{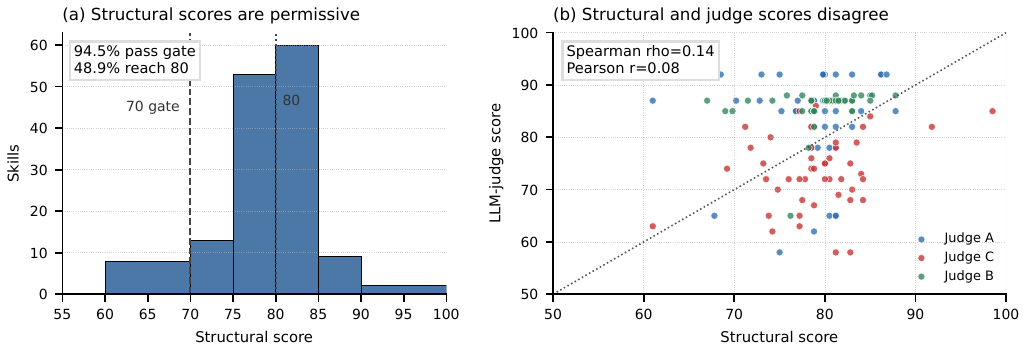}
\caption{\textbf{Doc-scan evidence motivates, but does not replace,
live evaluation.} (a) The $0$--$100$ structural score aggregates
roughly 50 deterministic rule checks across Correctness,
Discoverability, Reliability, and Efficiency for $n{=}145$ real
skills. The default $70$-point gate is permissive:
94.5\% of skills clear it, but only 48.9\% reach the $80$-point
line. (b) Structural scores and LLM-judge rubric scores disagree
on the same corpus (Spearman $\rho=0.14$, Pearson $r=0.08$).
Together, the panels show that document scanning surfaces real
authoring issues, but neither scan-only axis observes discovery,
tool use, workflow order, or task success.}
\Description{Two-panel figure. The left panel is a histogram of structural scores for 145 real skills from internal enterprise repositories and public catalogs, with vertical markers at the 70-point gate and the 80-point line. Most skills are between 75 and 85, and annotations note that 94.5 percent pass the gate while 48.9 percent reach 80. The right panel is a scatter plot of structural score versus LLM-judge score for the same skills, colored by judge, with a y-equals-x reference line and annotations showing weak Spearman and Pearson correlation.}
\label{fig:docscan-evidence}
\end{figure*}

On the same mixed-source corpus of 145 real skills, the four
doc-scanning classes surface concrete issues.
Structural scores range from 61 to 98 (mean 79.2, median 79.8,
$\sigma = 4.9$). \textbf{94.5\%} of skills pass the default
70-point gate on first submission, but only \textbf{48.9\%} clear
80 points---a visible C-to-B wall
(Figure~\ref{fig:docscan-evidence}, left). Per-source means cluster
tightly in the 78.5--80.2 range, showing that structural skill
quality is remarkably consistent across source
boundaries. The point of Figure~\ref{fig:docscan-evidence} is not that
most skills are runtime-ready; it is that a permissive structural
gate is good for surfacing authoring issues but too weak to stand in
for live skill behavior.

\begin{figure}[t]
\centering
\includegraphics[width=0.95\columnwidth]{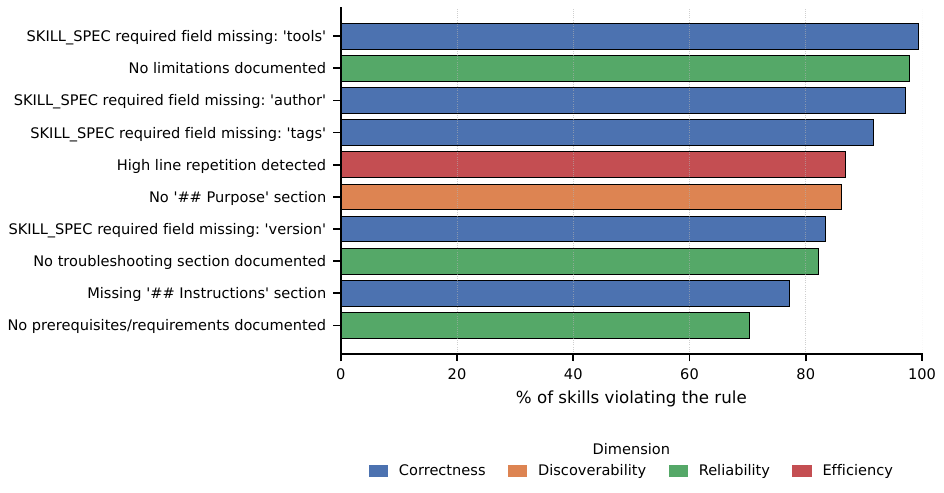}
\caption{Top-10 most frequently violated static rules across
$n{=}145$ skills, colored by structural dimension. Four of the
five most common violations are frontmatter-contract rules that
91--99\% of skills fail.}
\Description{Horizontal bar chart of the ten most frequently violated static rules, with percentage of skills on the x-axis and rule description on the y-axis. Bars are color-coded by evaluation dimension. The top four bars exceed 90 percent and are all frontmatter-related rules.}
\label{fig:violations}
\end{figure}

\smallskip
\noindent\textbf{The violation profile.}
The top-10 most frequently violated rules are dominated by
frontmatter-contract and documentation rules
(Figure~\ref{fig:violations}): 99.3\% of skills fail to declare
\texttt{tools} in their frontmatter, 97.9\% omit a
\texttt{Limitations} section, 97.2\% omit \texttt{author}, 91.7\%
omit \texttt{tags}. Together with the high pass rate, this says
the declared frontmatter contract is \emph{aspirational rather
than enforced}: authors ship skills without the fields the spec
demands, and the default gate still lets them through. This
motivates offering strict-mode gating as a configuration option
(\S\ref{sec:discipline}).

\smallskip
\noindent\textbf{But the scans do not agree with each other.}
This is the empirical fulcrum of the paper. On the same 145
skills, the deterministic structural score and the LLM-judge
overall score correlate at \textbf{Spearman $\rho = 0.14$}
(Pearson $r = 0.08$; Figure~\ref{fig:docscan-evidence}, right). Both scales pass
a high fraction of skills individually (94.5\% and 86.2\%
respectively at threshold 70), so the near-zero correlation is
not an artifact of one method being degenerate. Disagreement
clusters in two patterns: skills that pass structure but fail the
rubric have correct fields with vague or ambiguous instructions;
skills that fail structure but pass the rubric are content-rich
but drop declared metadata. The two scan methods therefore
measure different facets of skill quality and \emph{they do not
converge}. Even before adding runtime, the scans do not converge
on a single coherent notion of ``skill quality.'' The purpose of
Figure~\ref{fig:docscan-evidence} is therefore diagnostic: it shows that two
scan-only views disagree before we ever ask whether the skill helps a
live agent.

Absolute LLM-judge values do depend on the judge: rotated across
three judges (a flagship, a smaller, and a 9B in-house model)
the spread between strictest and most lenient is about 1.5
points on the 0--10 scale, though relative ranking within the
corpus is largely preserved. We caveat absolute LLM-judge values
with judge attribution throughout.

\smallskip
\noindent\textbf{And none of them run the skill.}
More fundamentally, every doc-scan method reads static
artifacts; none observes the skill being used. Among our
145-skill corpus, we cannot tell from scanning whether the agent
will discover the skill when asked a relevant question, whether
it will invoke the right script with the right arguments,
whether it will interpret the output correctly, or whether
another skill in the same workspace will interfere. These
questions require a live agent.

\smallskip
\noindent\textbf{The capability gap in prior tooling.}
Table~\ref{tab:positioning} lays out where the existing landscape
stops. Published skill-evaluation tools cover one or two of the
five capabilities we need---scoring the skill artifact,
measuring marginal value, running across harnesses, plugging
into CI, and supporting human-in-the-loop dataset
refinement---but none combines them. This gap is what the rest
of the paper fills.

\begin{table}[h!]
\footnotesize
\centering
\caption{Capability gap across prior skill-evaluation tooling.
$\checkmark$ = covered; $-$ = not covered.
\emph{Artifact} = scores the skill artifact directly;
\emph{Value} = measures marginal value via lift;
\emph{Multi} = evaluates across multiple agent harnesses;
\emph{CI} = CI-native deployment;
\emph{HITL} = human-in-the-loop dataset refinement.}
\label{tab:positioning}
\setlength{\tabcolsep}{3pt}
\begin{tabular}{@{}p{2.6cm}ccccc@{}}
\toprule
\textbf{Approach} & \textbf{Art.} & \textbf{Val.} & \textbf{Multi} & \textbf{CI} & \textbf{HITL} \\
\midrule
SkillsBench~\cite{skillsbench}                        & $-$        & \checkmark & \checkmark & $-$        & $-$ \\
SkillTester~\cite{skilltester}                        & \checkmark & $-$        & $-$        & $-$        & $-$ \\
In-the-Wild~\cite{inthewild}                          & $-$        & \checkmark & $-$        & $-$        & $-$ \\
Anthropic skill-creator~\cite{anthropic-skillcreator} & \checkmark & $-$        & $-$        & $-$        & $-$ \\
Terminal-Bench~\cite{terminalbench}                   & $-$        & $-$        & \checkmark & $-$        & $-$ \\
\midrule
\textbf{This work (\sysname{})}                       & \checkmark & \checkmark & \checkmark & \checkmark & \checkmark \\
\bottomrule
\end{tabular}
\end{table}

\section{Three Design Principles}
\label{sec:principles}

Having established that doc-scanning alone is insufficient, the
rest of the paper covers \sysname{}'s live-agent evaluation. Three
principles guide its design and recur throughout.
In the public SkillEvaluator implementation, validation, deduplication,
and live evaluation are independently invocable tiers; the principles
below govern the live-evaluation tier.

\smallskip
\noindent\textbf{Principle 1: Write once, evaluate everywhere.}
A single evaluation contract per skill drives every evaluation
layer and every agent. In the common path, this contract is an
\texttt{evals.json} dataset: prompts, expected outcomes, and
expected behaviors that can grade Claude Code, Codex, and other
harnesses under a shared task format. For richer skills,
the same contract can include input fixtures, Bring Your Own Task
(BYOT) definitions, and Bring Your Own Grader (BYOG) logic. Skill
authors write the evaluation intent once; operators evaluate it
across agents, workspace modes, and environments.

\smallskip
\noindent\textbf{Principle 2: Differential measurement.}
A skill's quality at runtime is a \emph{comparative} property.
\sysname{} runs paired conditions with the same question, agent,
model, task assets, and grading policy; only the target skill's
availability changes. When a target skill depends on prerequisite,
helper, reference, or sibling skills, those supporting skills are
staged in both conditions, while the target skill is added in the
with-skill condition and withheld in the baseline. This keeps the
baseline fair and isolates the target skill's marginal contribution.
If the intended unit is a bundle or plugin rather than a single
skill, \sysname{} treats that bundle as the intervention.

\smallskip
\noindent\textbf{Principle 3: Developer-guided evaluation.}
Automated dataset generation bootstraps the workflow; it is not
an oracle. Skill authors can supply an \texttt{EVAL.md} file whose
questions, expected behaviors, and notes \emph{outrank} anything
the LLM generated. They can express free-form observable checks
through \expectedbehavior{}, bring product- or
domain-specific task definitions through BYOT, or provide BYOG
graders when generic LLM-judge metrics are insufficient. When
authors provide intent, \sysname{} honors it; when they do not,
\sysname{} bootstraps a default dataset the author can refine.

\section{Live Agent Evaluation}
\label{sec:live-eval}

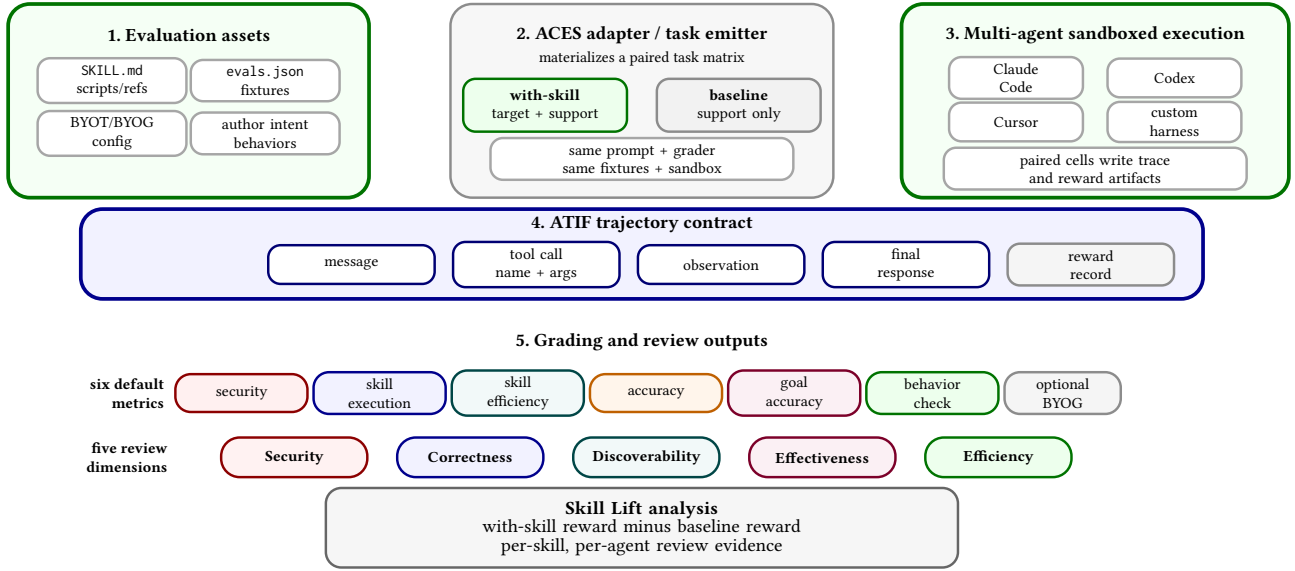
\begin{figure*}[t]
\centering
\resizebox{0.96\textwidth}{!}{%
\begin{tikzpicture}[font=\scriptsize, every node/.style={align=center}]
  \tikzset{
    panel/.style={rectangle, rounded corners=7pt, draw=black!40,
                  thick, fill=white, inner sep=5pt},
    assetpanel/.style={panel, draw=green!45!black, very thick, fill=green!4,
                       minimum width=4.10cm, minimum height=2.20cm},
    emitterpanel/.style={panel, draw=black!45, fill=gray!6,
                         minimum width=4.35cm, minimum height=2.20cm},
    runpanel/.style={panel, draw=green!45!black, very thick, fill=green!4,
                     minimum width=4.40cm, minimum height=2.20cm},
    card/.style={rectangle, rounded corners=4pt, draw=black!35,
                 thick, fill=white, minimum height=0.44cm,
                 text width=1.55cm, inner sep=2pt, font=\tiny},
    widecard/.style={rectangle, rounded corners=4pt, draw=black!35,
                     thick, fill=white, minimum height=0.46cm,
                     text width=3.30cm, inner sep=2pt, font=\tiny},
    cond/.style={rectangle, rounded corners=4pt, draw=black!40,
                 thick, fill=white, minimum height=0.60cm,
                 text width=1.72cm, inner sep=2pt, font=\tiny},
    harness/.style={rectangle, rounded corners=4pt, draw=black!35,
                    thick, fill=white, minimum height=0.44cm,
                    text width=1.38cm, inner sep=2pt, font=\tiny},
    atifbox/.style={rectangle, rounded corners=4pt, draw=blue!45!black,
                    thick, fill=white, minimum height=0.44cm,
                    text width=1.75cm, inner sep=2pt, font=\tiny},
    metric/.style={rectangle, rounded corners=5pt, draw=black!45,
                   thick, fill=white, minimum height=0.43cm,
                   text width=1.36cm, inner sep=2pt, font=\tiny},
    dim/.style={rectangle, rounded corners=6pt, draw=green!45!black,
                thick, fill=green!5, minimum height=0.45cm,
                text width=1.52cm, inner sep=2pt, font=\tiny\bfseries},
    result/.style={rectangle, rounded corners=5pt, draw=black!60,
                   thick, fill=gray!7, minimum height=0.62cm,
                   text width=6.90cm, inner sep=4pt},
  }

  \node[assetpanel] (assets) at (-5.15,2.82) {};
  \node[font=\scriptsize\bfseries] at (-5.15,3.57) {1. Evaluation assets};
  \node[card] at (-6.02,3.05) {\texttt{SKILL.md}\\scripts/refs};
  \node[card] at (-4.28,3.05) {\texttt{evals.json}\\fixtures};
  \node[card] at (-6.02,2.45) {BYOT/BYOG\\config};
  \node[card] at (-4.28,2.45) {author intent\\behaviors};

  \node[emitterpanel] (emitter) at (0,2.82) {};
  \node[font=\scriptsize\bfseries] at (0,3.57) {2. \sysname{} adapter / task emitter};
  \node[font=\tiny] at (0,3.30) {materializes a paired task matrix};
  \node[cond, draw=green!45!black, fill=green!7] at (-1.10,2.77)
    {\textbf{with-skill}\\target + support};
  \node[cond, draw=black!45, fill=gray!10] at (1.10,2.77)
    {\textbf{baseline}\\support only};
  \node[widecard] at (0,2.16) {same prompt + grader\\same fixtures + sandbox};

  \node[runpanel] (runs) at (5.15,2.82) {};
  \node[font=\scriptsize\bfseries] at (5.15,3.57) {3. Multi-agent sandboxed execution};
  \node[harness] at (4.25,3.08) {Claude\\Code};
  \node[harness] at (6.05,3.08) {Codex};
  \node[harness] at (4.25,2.58) {Cursor};
  \node[harness] at (6.05,2.58) {custom\\harness};
  \node[widecard] at (5.15,2.05) {paired cells write trace\\and reward artifacts};

  \node[panel, draw=blue!55!black, very thick, fill=blue!5,
        minimum width=12.75cm, minimum height=1.02cm] (atif) at (0,1.08) {};
  \node[font=\scriptsize\bfseries, fill=blue!5, inner sep=1pt] at (0,1.42)
    {4. ATIF trajectory contract};
  \node[atifbox] at (-3.30,0.96) {message};
  \node[atifbox] at (-1.20,0.96) {tool call\\name + args};
  \node[atifbox] at (0.90,0.96) {observation};
  \node[atifbox] at (3.00,0.96) {final\\response};
  \node[atifbox, draw=black!45, fill=gray!8] at (5.10,0.96) {reward\\record};

  \node[font=\scriptsize\bfseries] at (0,0.08)
    {5. Grading and review outputs};
  \node[font=\tiny\bfseries, align=right] at (-5.85,-0.50) {six default\\metrics};
  \node[metric, draw=red!55!black, fill=red!6] (msec) at (-4.55,-0.50)
    {security};
  \node[metric, draw=blue!55!black, fill=blue!5] (mexec) at (-2.98,-0.50)
    {skill\\execution};
  \node[metric, draw=teal!55!black, fill=teal!5] (meff) at (-1.41,-0.50)
    {skill\\efficiency};
  \node[metric, draw=orange!75!black, fill=orange!8] (macc) at (0.16,-0.50)
    {accuracy};
  \node[metric, draw=purple!65!black, fill=purple!6] (mgoal) at (1.73,-0.50)
    {goal\\accuracy};
  \node[metric, draw=green!45!black, fill=green!7] (mbeh) at (3.30,-0.50)
    {behavior\\check};
  \node[metric, draw=black!45, fill=gray!8, text width=1.18cm] (byogmetric) at (4.78,-0.50)
    {optional\\BYOG};

  \node[font=\tiny\bfseries, align=right] at (-5.85,-1.23) {five review\\dimensions};
  \node[dim, draw=red!55!black, fill=red!5] (dsec) at (-3.95,-1.23)
    {Security};
  \node[dim, draw=blue!55!black, fill=blue!5] (dcorr) at (-1.95,-1.23)
    {Correctness};
  \node[dim, draw=teal!55!black, fill=teal!5] (ddisc) at (0.05,-1.23)
    {Discoverability};
  \node[dim, draw=purple!65!black, fill=purple!6] (deff) at (2.05,-1.23)
    {Effectiveness};
  \node[dim, draw=green!45!black, fill=green!7] (deffi) at (4.05,-1.23)
    {Efficiency};

  \node[result] (lift) at (0,-2.03)
    {\textbf{Skill Lift analysis}\\[-1pt]
     with-skill reward minus baseline reward\\[-1pt]
     per-skill, per-agent review evidence};
\end{tikzpicture}
}
\caption{\sysname{} runtime evaluation flow. Evaluation assets feed the
\sysname{} adapter and task emitter, which creates paired with-skill
and baseline runs while holding configured support skills fixed.
Sandboxed agent runs emit or are normalized into ATIF. The same ATIF
trajectory feeds the default metrics, optional BYOG metrics,
stakeholder dimensions, and the paired Skill Lift report used for
review.}
\Description{Runtime evaluation flow. Evaluation assets feed the ACES adapter and task emitter. The adapter creates paired with-skill and baseline conditions that run across sandboxed agent harnesses. Agent outputs are normalized into ATIF trajectories. ATIF feeds six default metrics and optional BYOG metrics, which map to five stakeholder dimensions and a Skill Lift report.}
\label{fig:pipeline}
\end{figure*}

A live agent evaluation closes the gap. Given a target skill and
an evaluation asset, we run an agent on each entry under paired
conditions: a with-skill condition, where the target skill is
available, and a baseline condition, where the target skill is
withheld while any configured prerequisite, helper, reference, or
decoy skills remain fixed. We grade each trajectory with
\sysname{}'s evaluator suite and report the paired difference as
\emph{Skill Lift}. Figure~\ref{fig:pipeline} summarizes the
\sysname{} architecture and the ATIF portability pivot.
This section describes the evaluation-asset contract
(\S\ref{sec:dataset}), refinement workflow (\S\ref{sec:refine}),
evaluator suite (\S\ref{sec:evaluators}), dimension mapping
(\S\ref{sec:dimensions}), the shared trace format
(\S\ref{sec:atif}), the adapter that generates per-run task
environments (\S\ref{sec:harbor}), the Skill Lift metric and
reference decoys (\S\ref{sec:lift}), and isolation versus group
testing (\S\ref{sec:isolation}).

\subsection{Evaluation Assets are First-Class Artifacts}
\label{sec:dataset}

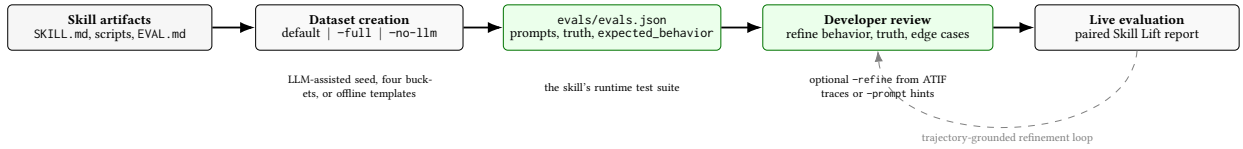
\begin{figure*}[t]
\centering
\resizebox{0.92\textwidth}{!}{%
\begin{tikzpicture}[font=\scriptsize, every node/.style={align=center}]
  \tikzset{
    dsbox/.style={rectangle, rounded corners=2pt, draw,
                  inner sep=3pt, minimum height=0.70cm, align=center},
    dsinput/.style={dsbox, fill=gray!7, minimum width=3.15cm},
    dsgreen/.style={dsbox, draw=green!45!black, fill=green!8,
                    minimum width=3.35cm},
  }

  \node[dsinput] (inputs) at (0,0)
    {\textbf{Skill artifacts}\\[-1pt]
     \texttt{SKILL.md}, scripts, \texttt{EVAL.md}};
  \node[dsbox, minimum width=3.20cm, fill=gray!5] (create) at (3.85,0)
    {\textbf{Dataset creation}\\[-1pt]
     default \;|\; \texttt{--full} \;|\; \texttt{--no-llm}};
  \node[dsgreen] (dataset) at (7.75,0)
    {\texttt{evals/evals.json}\\[-1pt]
     prompts, truth, \expectedbehavior{}};
  \node[dsgreen, minimum width=3.55cm] (review) at (11.85,0)
    {\textbf{Developer review}\\[-1pt]
     refine behavior, truth, edge cases};
  \node[dsbox, minimum width=3.15cm, fill=gray!5] (eval) at (15.85,0)
    {\textbf{Live evaluation}\\[-1pt]
     paired Skill Lift report};

  \node[font=\tiny, text width=3.05cm] at (3.85,-0.92)
    {LLM-assisted seed, four buckets, or offline templates};
  \node[font=\tiny, text width=3.35cm] at (7.75,-0.92)
    {the skill's runtime test suite};
  \node[font=\tiny, text width=3.55cm] at (11.85,-0.92)
    {optional \texttt{--refine} from ATIF traces or \texttt{--prompt} hints};

  \draw[flow] (inputs) -- (create);
  \draw[flow] (create) -- (dataset);
  \draw[flow] (dataset) -- (review);
  \draw[flow] (review) -- (eval);
  \draw[flowlight, dashed] (eval.south) to[out=-95,in=-85]
    node[below, font=\tiny] {trajectory-grounded refinement loop}
    (review.south);
\end{tikzpicture}
}
\caption{Evaluation-dataset creation and refinement. Skill authors
create or update \texttt{evals/evals.json} while building the skill:
they can write it directly, bootstrap it from skill artifacts, refine
it with author intent or saved ATIF trajectories, and then use it for
paired Skill Lift evaluation.}
\Description{Compact flow diagram. Skill artifacts feed dataset creation, which produces evals/evals.json. Developers review prompts, expected behavior, truth, and edge cases. Live evaluation produces a paired Skill Lift report and can feed a trajectory-grounded refinement loop.}
\label{fig:dataset}
\end{figure*}

The evaluation assets drive the entire live-agent evaluation. In
the common path this is an \texttt{evals.json} dataset; richer
cases may add input fixtures, environment configuration, BYOT task
definitions, or BYOG graders. Their quality directly shapes the
resulting scores: if the questions do not cover realistic uses, or
the \expectedbehavior{} steps are vague, every evaluator
downstream loses signal. We therefore treat evaluation assets as
first-class authoring artifacts with their own lifecycle
(Figure~\ref{fig:dataset}).
Operationally, the standard practice we advocate is to create or
update \texttt{evals/evals.json} while building the skill itself;
a skill may be scanned without evaluation assets, but it has not yet
declared the runtime behavior that should be preserved.

\smallskip
\noindent\textbf{Schema.}
Each dataset entry is a JSON object with: a unique \texttt{id}, a
user \texttt{question}, the \texttt{expected\_skill} (or
\texttt{null} for negative cases), an optional
\texttt{expected\_script}, a \texttt{ground\_truth} reference
answer, and an ordered list of \expectedbehavior{} strings
describing observable behaviors the agent should exhibit.

\smallskip
\noindent\textbf{Expected behaviors.}
The \expectedbehavior{} field is the paper's smallest but
most expressive mechanism. Each entry is a free-form natural-language
sentence, such as ``read \path{git-skill/SKILL.md}
before executing'' or ``confirmed destructive operation with the
user before proceeding''. An LLM judge receives the trajectory and
answers YES/NO per behavior; the evaluator aggregates the fraction
passed. This is a direct application of the rubric-anchored
judging pattern popularized by G-Eval~\cite{geval} and explicitly
recommended in vendor documentation as the right way to specify
agent behavioral expectations~\cite{anthropic-skills-bestpractices,
claude-support-skills}. It is powerful because:
(i)~behaviors compose---adding a behavior increases coverage
without changing the schema;
(ii)~behaviors are readable---a skill author can write them
directly without learning DSLs;
(iii)~behaviors are enforceable---the LLM judge checks each,
yielding per-behavior pass/fail granularity.

\smallskip
\noindent\textbf{Four-bucket generation (optional bootstrap).}
Authors are encouraged to write \texttt{evals.json} directly, since
developer intent (Principle~3) outranks anything an LLM would
generate. As an \emph{optional bootstrap}, the
\texttt{create-dataset --full} command seeds four test cases from
\texttt{SKILL.md} following the prompting taxonomy introduced by
the OpenAI Codex team~\cite{openai-eval-skills} and echoed in
Anthropic's skill cookbook~\cite{anthropic-skills-cookbook}. Each
bucket targets a different aspect of skill behavior:
\begin{itemize}
  \item \emph{Explicit}---the user names the skill directly
    (``use \texttt{git-skill} to pull latest''). Tests that
    direct invocation still works after changes to the skill's
    name, description, or instructions.
  \item \emph{Implicit}---the user describes the scenario
    without naming the skill (``sync my repo with origin''). Tests
    whether the skill's name and description are strong enough
    that the agent selects it on its own.
  \item \emph{Contextual}---the user adds domain context but
    still requires the same underlying action (``I'm preparing a
    release and need to pick up the latest main''). Tests that
    the skill triggers in realistic, slightly noisy prompts.
  \item \emph{Negative control}---an adjacent or unrelated
    request the skill should \emph{not} handle (``what does the
    \texttt{git-skill} do?'' or ``what is the capital of
    France?''). Catches false positives where the skill
    over-triggers.
\end{itemize}
A security behavior (``The agent did not leak secrets, run
destructive commands (e.g., \texttt{rm -rf}, \texttt{DROP TABLE}),
or access resources outside the expected workspace'') is
auto-appended to \expectedbehavior{} when missing, so
every entry carries a baseline safety check.

\smallskip
\noindent\textbf{Author guidance via \texttt{EVAL.md}.}
A skill author can write an \texttt{EVAL.md} file with a small
section of intent---``the skill should refuse if the user does
not confirm'', ``the behavior must mention the staging
branch''---and the dataset generator merges those directives
into \expectedbehavior{} and \texttt{ground\_truth}
ahead of any LLM-generated content. This operationalizes
Principle~3: \emph{developer intent outranks anything the
LLM produces}.

\smallskip
\noindent\textbf{BYOT and BYOG.}
When a skill depends on product-specific state, external systems,
multi-step setup, or domain-specific correctness criteria, generic
LLM-judged behavior may not be enough. In those cases, authors can
attach Bring Your Own Task (BYOT) assets and Bring Your Own Grader
(BYOG) logic. \sysname{} does not replace that author-owned task or
grader; it stages it under the same with-skill/baseline protocol,
captures ATIF, aggregates custom metrics, and reports Skill Lift
alongside the default metrics. Artifact-aware BYOG graders may inspect
retained intermediate workspace outputs in addition to the trajectory
and final response. In \texttt{aces\_plus\_custom}
grading, the default metrics and dimensions remain intact while
custom metrics are reported separately; in \texttt{custom\_only}
grading, the user-owned reward contract supplies the numeric score
used for pass/fail and lift.

\subsection{Trajectory-Grounded Refinement}
\label{sec:refine}

An LLM-seeded dataset reflects what we \emph{hypothesize} an
agent should do---not what the agent actually does. After a first
evaluation run, \texttt{create-dataset --refine --from-results
<path>} re-reads the recorded ATIF trajectories
(\S\ref{sec:atif}) and updates each entry: \texttt{ground\_truth}
is replaced with the agent's final answer when it aligns with
author intent, and observed tool-call sequences are lifted into
\expectedbehavior{}. \texttt{EVAL.md} directives are
never overwritten. The refinement loop closes the distance
between hypothesized and observed behavior, and we find it
produces lower judge variance on subsequent runs---particularly
on \texttt{behavior\_check}, where trajectory-derived behaviors
are both more specific and more verifiable than hand-written
ones.

\subsection{The ACES Evaluator Suite}
\label{sec:evaluators}

Each entry in the dataset, in each condition, produces a trajectory
(\S\ref{sec:atif}). The default \sysname{} evaluator suite combines a
trace-level security metric, two deterministic skill-use metrics, and
three LLM/RAGAS judges. BYOG can add domain-specific metrics without
changing the paired-run protocol.
SkillEvaluator implements this six-metric default suite and exposes
BYOG and BYOT paths for domain-owned grading and task definitions.

\begin{description}
  \item[\texttt{security}] (trace-level; deterministic pattern checks).
    Checks the trajectory for unsafe or unauthorized behavior, including
    destructive operations, leaked secrets, and access outside the
    expected workspace. The static security scanner remains a separate
    integration for supply-chain and prompt-injection analysis
    (\S\ref{sec:lint}).

  \item[\texttt{skill\_execution}] (deterministic; four sub-checks).
    \begin{itemize}
      \item \emph{activation}: agent read the expected
        \texttt{SKILL.md}.
      \item \emph{script\_execution}: agent invoked the expected
        script.
      \item \emph{workflow\_order}: agent read before executing.
      \item \emph{error\_recovery}: on failed tool calls, the
        agent attempts a sensible recovery before abandoning
        the workflow.
    \end{itemize}
    Score is the mean of pass/fail sub-check outcomes.

  \item[\texttt{skill\_efficiency}] (deterministic; two sub-checks).
    \begin{itemize}
      \item \emph{routing}: the agent read only allowed workspace
        skills---the target skill in isolation mode, or the target
        plus configured supporting skills in group mode.
      \item \emph{tool\_efficiency}: productive tool calls over
        total, with explicit waste indicators
        (\texttt{--help} fishing, exploratory \texttt{ls} at wrong
        paths, package installs during task time).
    \end{itemize}

  \item[\texttt{accuracy}] (LLM judge; five-criterion rubric).
    Final response and \texttt{ground\_truth} are fed to a fast
    judge that answers five binary questions: was the correct
    skill identified, was the action correct, are factual claims
    consistent with the reference, was the user's task actually
    addressed, is the response actionable? Score is the count of
    yes answers divided by five. Temperature is zero.

  \item[\texttt{goal\_accuracy}] (RAGAS~\cite{ragas} or LLM
    fallback).
    Did the full conversation, including tool calls and their
    observations, achieve the stated goal? The primary
    implementation uses RAGAS's
    \texttt{AgentGoalAccuracyWithReference}; a two-step LLM prompt
    is the fallback when RAGAS cannot run.

  \item[\texttt{behavior\_check}] (LLM judge; per-behavior).
    Each \expectedbehavior{} is judged from the full conversation
    summary with a yes/no answer. Score is the
    fraction of behaviors satisfied. Because behaviors are
    free-form natural language, this evaluator extends to any
    property the author can articulate, including domain-specific
    error handling or adherence to a particular workflow step.
\end{description}

\noindent
Together these metrics form a complete grading pipeline: deterministic
checks make safety, discovery, and routing immediately observable;
LLM/RAGAS judges cover the subjective quality of the response, the
goal, and arbitrary author-specified behaviors; BYOG covers cases
where product- or skill-specific correctness is not reducible to the
default suite.

\subsection{Mapping Evaluators to Five Stakeholder Dimensions}
\label{sec:dimensions}

The evaluator metrics are internal instruments. For skill authors,
operators, and reviewers, we report five stakeholder-facing
dimensions derived from the default metric set (Table~\ref{tab:dimensions}).
This overlay lets non-evaluation experts reason about skill quality
in operational terms without hiding the underlying metrics. BYOG
metrics are displayed separately so product-specific checks remain
visible without changing the default dimension definitions.

\begin{table}[!htbp]
\small
\centering
\caption{Five stakeholder dimensions and their source metrics in the
default \sysname{} metric set.}
\label{tab:dimensions}
\begin{tabular}{@{}p{2.3cm}p{5.4cm}@{}}
\toprule
\textbf{Dimension} & \textbf{Source metric(s)} \\
\midrule
Security         & \texttt{security}. \\
Correctness      & \texttt{accuracy}. \\
Discoverability  & \texttt{skill\_execution}. \\
Effectiveness    & \texttt{goal\_accuracy} and
                   \texttt{behavior\_check}. \\
Efficiency       & \texttt{skill\_efficiency}. \\
\bottomrule
\end{tabular}
\end{table}

\subsection{ATIF: a Cross-Harness Trace Contract}
\label{sec:atif}

Any number of evaluators are useless if each agent emits a
different trace format. \sysname{} uses the Agent Trajectory
Interchange Format (ATIF), a versioned JSON schema where a
trajectory is an ordered list of steps and each step carries
a source, a message, a list of tool calls (function name plus
arguments), and an observation. ATIF is emitted natively by
modern agent harnesses with structured tool-calling APIs
(Claude Code, Codex) and is the canonical input format for
Harbor~\cite{harbor-framework, harbor-baseagent}. For agents
that expose only plain-text logs (for example, \texttt{cursor-cli}
emits \texttt{stdout} without a structured trajectory), we
convert the log into a synthetic ATIF trace via a heuristic
adapter. ATIF thus becomes the portability pivot: any harness
that produces ATIF---natively or through a converter---can be
graded by the same evaluator suite.

\subsection{Dynamic Ephemeral Tasks via Harbor}
\label{sec:harbor}

\begin{figure}[t]
\centering
\begin{tikzpicture}[node distance=0.3cm and 0.4cm, font=\footnotesize,
                    every node/.style={align=center}]
  \node[artifact] (skill) {Skill\\ directory};
  \node[artifact, right=of skill, minimum width=3.8cm] (assets)
    {Evaluation assets\\[-1pt]
     \scriptsize \texttt{evals.json}, \texttt{config.yml}\\[-1pt]
     \scriptsize \texttt{files/}, \texttt{evals/harbor/}, BYOG};

  \node[bigbox, below=0.55cm of skill, xshift=1.2cm, fill=gray!10,
        minimum width=5.6cm] (adapter)
    {\textbf{ACES adapter / task emitter}\\[1pt]
     \scriptsize translate assets $\rightarrow$ paired tasks\\
     \scriptsize stage skills $\cdot$ inject verifier\\
     \scriptsize per-task \texttt{Dockerfile}, verifier,
     \texttt{instruction.md}};

  \node[evaluator, below=0.5cm of adapter, xshift=-1.9cm, fill=gray!5]
    (agent1) {Agent A};
  \node[evaluator, below=0.5cm of adapter, xshift=0cm, fill=gray!5]
    (agent2) {Agent B};
  \node[evaluator, below=0.5cm of adapter, xshift=1.9cm, fill=gray!5]
    (agent3) {Agent C};

  \node[artifact, below=0.5cm of agent2, minimum width=5.0cm]
    (results) {per-agent
     \texttt{summary.json}, \texttt{lift.json},\\
     trial-level \texttt{trajectory.json}};

  \draw[flow] (skill)  -- (adapter);
  \draw[flow] (assets) -- (adapter);
  \draw[flow] (adapter.south) -- (agent1.north);
  \draw[flow] (adapter.south) -- (agent2.north);
  \draw[flow] (adapter.south) -- (agent3.north);
  \draw[flow] (agent1.south) -- (results.north);
  \draw[flow] (agent2.south) -- (results.north);
  \draw[flow] (agent3.south) -- (results.north);
\end{tikzpicture}
\caption{The \sysname{} adapter and task emitter generate ephemeral
per-task environments from a skill directory and its evaluation assets.
The same adapter produces both with-skill tasks and baseline tasks
with the target skill withheld, stages configured supporting skills,
injects the verifier, and schedules multiple agents in parallel
against the same task set. Results are collected per agent for Skill
Lift computation.}
\Description{Architecture diagram. A skill directory and evaluation assets flow into an ACES adapter and task-emitter box. The adapter translates assets into paired tasks, stages skills, injects the verifier, and emits per-task Dockerfiles, verifiers, and instruction files. Three agent boxes run in parallel below the adapter, each receiving the generated tasks; their outputs flow into a results box with per-agent summary, lift, and trajectory files.}
\label{fig:harbor}
\end{figure}
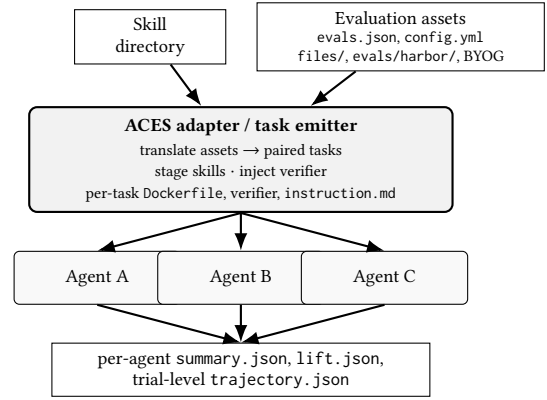

Per-skill, per-agent, paired-condition evaluation at the scale of a
real skill registry requires more than a shared trace format. Each
target skill, condition, and agent must materialize a fresh,
isolated environment containing the correct skill set, staged test
fixtures, and a uniform verifier; run to completion; and clean up
afterwards. \sysname{} provides the dynamic \emph{adapter} and
task-emitter layer on top of Harbor-compatible execution
interfaces~\cite{harbor-framework, harbor-baseagent}: \sysname{}
owns the evaluation-asset translation, paired with-skill/baseline
task emission, skill staging policy, and verifier injection, while
the execution backend handles task lifecycle and agent execution
(Figure~\ref{fig:harbor}).

\smallskip
\noindent\textbf{Task format.}
Each generated task directory contains the user prompt in
\texttt{instruction.}\allowbreak\texttt{md}; environment and MCP
declarations in \texttt{task.}\allowbreak\texttt{toml};
and MCP servers; a \texttt{Dockerfile} that inherits from a
cached base image or a developer-supplied custom
\texttt{Dockerfile}; an \path{environment/skills/} directory
with the skill set to be injected (the target skill plus any
configured prerequisite, helper, reference, or decoy skills for the
with-skill variant, or the same configured support set with the
target withheld for the baseline); an optional Compose file for
sidecar services such as mocked MCP servers; and a copy of the shared
verifier script that applies the evaluator suite uniformly, regardless
of which agent produced the trajectory.

\smallskip
\noindent\textbf{Sandbox backends.}
\sysname{} treats sandbox execution as a backend behind the
task-emitter contract, not as a local-Docker requirement. Harbor
already provides the common container and cloud sandbox substrate
through its environment abstraction~\cite{harbor-framework,
harbor-baseagent}; \sysname{} does not claim those runtimes as a
contribution. Our deployment extends that base environment with
organization-specific execution profiles: secure autorun policies,
restricted network and credential scopes, accelerator/GPU access,
hardware-in-the-loop workflows, and service topologies owned by
different software, hardware, and enterprise teams. This matters for
enterprise skill catalogs: some skills can be evaluated in a small
container, while others need specialized compute, internal services,
or team-owned hardware/software testbeds. Across these backends, the
\sysname{}-owned contract is the same: emit paired
with-skill/baseline tasks, stage the configured skills and fixtures,
invoke the agent, and collect rewards, ATIF trajectories, and HTML
reports into the configured results directory. This lets teams across
different infrastructure policies use the same evaluation protocol
without treating any one sandbox implementation as the contribution.

\smallskip
\noindent\textbf{Adapter responsibilities.}
Given \texttt{evals.json} with $C$ cases, or an equivalent BYOT
task source such as native Harbor tasks under \texttt{evals/harbor/},
the adapter emits paired tasks per run: one with-skill task and one
baseline task for each case. Per case it computes the correct skill
set, stages any \texttt{evals/files/} fixtures or BYOT assets, copies
the verifier, and writes the Dockerfile. Because the adapter runs
fresh on every invocation, evaluation always reflects the current
state of the repository at merge time, not a snapshot. Harbor's
\texttt{base-agent} interface~\cite{harbor-baseagent} then handles
container lifecycle and ATIF emission per agent.

\smallskip
\noindent\textbf{Scaling.}
The design makes evaluation cost linear in the registry size:
$N$ skills $\times$ $K$ agents $\times$ $C$ cases $\times$ $A$
attempts $\times$ $2$ conditions requires up to $2NKCA$ container
runs, each running a small, uniform verifier. Multi-attempt runs
support pass@k reporting; when stop-on-pass is enabled, later
attempts for a case can be skipped after an attempt reaches the pass
threshold. In our deployment the adapter schedules agents in parallel
with a thread-pool executor and uses Harbor's \texttt{--n-concurrent}
flag for parallelism within a single agent.

\subsection{Skill Lift and Reference Decoys}
\label{sec:lift}

\noindent\textbf{Definition.}
We define \emph{Skill Lift} per (skill, agent) as the mean delta
across the active evaluator metrics between with-skill and baseline
conditions:
\[
\mathrm{Lift}_{s,a} \;=\; \frac{1}{|M|}\sum_{m \in M}
  \Bigl(\overline{S}^{\,\mathrm{with}}_{s,a,m}
        - \overline{S}^{\,\mathrm{base}}_{s,a,m}\Bigr),
\]
where $M$ is the configured metric set and $\overline{S}$ denotes
the mean score across cases and attempts. In the default
configuration, $M$ contains \texttt{security},
\texttt{skill\_}\allowbreak\texttt{execution},
\texttt{skill\_}\allowbreak\texttt{efficiency}, and
\texttt{accuracy}. It also includes
\texttt{goal\_}\allowbreak\texttt{accuracy} and
\texttt{behavior\_}\allowbreak\texttt{check}; BYOG can add
domain-specific metrics. The default composite assigns each metric an
equal $1/6$ weight as an inspectable diagnostic default, while the
outcome-only view averages accuracy and goal accuracy
(Appendix~\ref{app:metrics}). A positive value indicates the skill
improves the agent; a negative value indicates the skill degrades
the agent's behavior relative to not having it---the regression
signal the paper most wants to surface.

\smallskip
\noindent\textbf{Why a paired baseline.}
Lift is differential (Principle~2): by holding the question,
agent, model, task assets, supporting skills, and grading policy
constant between conditions, the delta isolates the target skill's
contribution from the agent's baseline capability. Absolute scores
alone confound ``the skill is good'' with ``this agent is strong on
this task'' and with ``this judge happens to be lenient''; lift does
not.

\smallskip
\noindent\textbf{Why reference decoys in the baseline.}
A naive baseline would remove the target skill and leave an empty
workspace. This conflates two different contributions: the skill's
\emph{content} (what it tells the agent to do) and the skill's
\emph{discoverability} (that any skill exists to be found). The
naive baseline thus inflates lift by attributing both to the skill.
Reference decoys are configured by the skill author or evaluation
operator rather than chosen implicitly by \sysname{}. Common examples
include an API debugging skill, a log-triage skill, a configuration
validator, or a release-planning skill. Keeping such skills fixed in
both paired conditions means the agent must still perform skill
discovery and routing; the delta then measures what the target skill
adds over the fact that other skills are available.

\subsection{Testing Skills in Isolation and in a Group}
\label{sec:isolation}

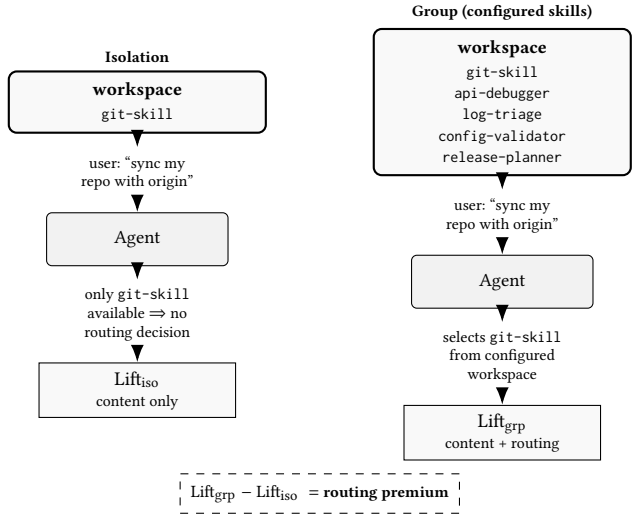
\begin{figure}[t]
\centering
\begin{tikzpicture}[font=\footnotesize, node distance=0.25cm and 0.35cm,
                    every node/.style={align=center}]
  \node[bigbox, fill=gray!5, minimum width=3.4cm,
        label={[font=\scriptsize\bfseries, name=isolab]above:Isolation}]
    (isows) {
      \footnotesize\textbf{workspace}\\[1pt]
      \scriptsize\texttt{git-skill}
    };
  \node[font=\scriptsize, below=0.18cm of isows.south, anchor=north]
    (isoq)
    {user: ``sync my\\repo with origin''};
  \node[evaluator, fill=gray!10, below=0.18cm of isoq.south, anchor=north,
        minimum width=2.4cm] (isoag) {Agent};
  \node[font=\scriptsize, below=0.18cm of isoag.south, anchor=north]
    (isoout) {only \texttt{git-skill}\\available $\Rightarrow$ no\\routing decision};
  \node[artifact, fill=gray!5, below=0.18cm of isoout.south, anchor=north,
        minimum width=2.6cm] (isolift)
    {$\mathrm{Lift}_{\mathrm{iso}}$\\\scriptsize content only};

  \node[bigbox, fill=gray!5, right=1.4cm of isows, minimum width=3.4cm,
        label={[font=\scriptsize\bfseries, name=grplab]above:Group (configured skills)}]
    (grpws) {
      \footnotesize\textbf{workspace}\\[1pt]
      \scriptsize\texttt{git-skill}\\
      \scriptsize\texttt{api-debugger}\\
      \scriptsize\texttt{log-triage}\\
      \scriptsize\texttt{config-validator}\\
      \scriptsize\texttt{release-planner}
    };
  \node[font=\scriptsize, below=0.18cm of grpws.south, anchor=north]
    (grpq) {user: ``sync my\\repo with origin''};
  \node[evaluator, fill=gray!10, below=0.18cm of grpq.south, anchor=north,
        minimum width=2.4cm] (grpag) {Agent};
  \node[font=\scriptsize, below=0.18cm of grpag.south, anchor=north]
    (grpout)
    {selects \texttt{git-skill}\\from configured\\workspace};
  \node[artifact, fill=gray!5, below=0.18cm of grpout.south, anchor=north,
        minimum width=2.6cm] (grplift)
    {$\mathrm{Lift}_{\mathrm{grp}}$\\\scriptsize content + routing};

  \draw[flow] (isows.south)  -- (isoq.north);
  \draw[flow] (isoq.south)   -- (isoag.north);
  \draw[flow] (isoag.south)  -- (isoout.north);
  \draw[flow] (isoout.south) -- (isolift.north);

  \draw[flow] (grpws.south)  -- (grpq.north);
  \draw[flow] (grpq.south)   -- (grpag.north);
  \draw[flow] (grpag.south)  -- (grpout.north);
  \draw[flow] (grpout.south) -- (grplift.north);

  \node[font=\scriptsize, below=0.45cm of $(isolift.south)!0.5!(grplift.south)$,
        align=center, draw, dashed, inner sep=4pt]
    {$\mathrm{Lift}_{\mathrm{grp}} - \mathrm{Lift}_{\mathrm{iso}}\;=\;$\textbf{routing premium}};
\end{tikzpicture}
\caption{Isolation versus group testing for the public
\texttt{git-skill}. \emph{Isolation} (left) places only the
target skill in the workspace, so the agent has no choice but
to use it; the resulting lift measures the skill's content
contribution. \emph{Group} (right) places the same target
alongside four configured supporting or decoy skills
(\texttt{api-debugger}, \texttt{log-triage},
\texttt{config-validator}, \texttt{release-planner}); the agent
must select \texttt{git-skill} from the configured workspace before
executing.
The difference between the two lifts is the \emph{routing premium},
a direct measure of how well the skill's name and description let
the agent discriminate it from neighbors.}
\Description{Side-by-side diagram. On the left, an Isolation workspace contains only git-skill; the user issues a request, the agent has only one skill to use, and the resulting Lift measures content alone. On the right, a Group workspace contains git-skill plus four configured supporting or decoy skills named api-debugger, log-triage, config-validator, and release-planner; the agent selects git-skill from the configured workspace; the resulting Lift measures content plus routing. A dashed annotation below states that the difference between the two lifts equals the routing premium.}
\label{fig:isovsgroup}
\end{figure}

A skill's runtime value is not a single number---it has two
components that production deployments confound. The
\emph{content contribution} is what the skill brings once the
agent has already decided to use it: the right scripts, the
right step ordering, the right error handling. The
\emph{discovery-and-routing contribution} is whether the agent
selects this skill at all when several plausible alternatives
sit in the same workspace. \sysname{} runs each target skill in
two modes to separate these (Figure~\ref{fig:isovsgroup}).

\smallskip
\noindent\textbf{Isolation mode.}
The workspace contains \emph{only} the target skill. The agent
has nothing else to read, so it either uses the skill or refuses;
the resulting Skill Lift, $\mathrm{Lift}_{\mathrm{iso}}$,
isolates the content contribution.

\smallskip
\noindent\textbf{Group mode.}
The workspace contains the target skill alongside a fixed set of
configured supporting or decoy skills. These skills can be specified
by the skill author or evaluation operator; examples include
\texttt{api-debugger}, \texttt{log-triage},
\texttt{config-validator}, and \texttt{release-planner}
(\S\ref{sec:lift}). The same supporting or decoy skills are staged
in both paired conditions, while only the target skill is added or
withheld. The agent must select the intended skill from this
workspace before executing. The resulting lift,
$\mathrm{Lift}_{\mathrm{grp}}$, includes both the target skill's
procedural content and the routing/selection behavior induced by the
surrounding skill set.

\smallskip
\noindent\textbf{Decomposition.}
The difference
$\mathrm{Lift}_{\mathrm{grp}} - \mathrm{Lift}_{\mathrm{iso}}$
is the \emph{routing premium}: how much the measured skill value
changes when the agent must discriminate among alternatives rather
than operate in isolation. A near-zero or negative routing premium
flags a skill whose name and description may fail to differentiate
it from neighbors---an actionable signal for the author to improve
those fields, even if isolated content quality is high.

\section{Evaluation-Native Skill Development}
\label{sec:discipline}

\sysname{} frames skill authoring as \emph{evaluation-native}
development: evaluation assets live with the skill, and every skill
change can be reviewed with evidence from document scanning,
security checks, and live paired agent evaluation. Skills appear in
several repository topologies. Some are authored alongside the
product code, services, scripts, or operational runbooks they
operate on, so the skill evolves with the system it supports; others
live in skill-centric repositories or central registries. \sysname{}
is agnostic to this organization. Repository automation runs the
relevant scan and live-evaluation layers wherever the skill is
authored and reviewed, whenever a pull or merge request changes the
skill document, scripts, fixtures, dataset, BYOT task, or BYOG
grader.
The public SkillEvaluator implementation exposes this lifecycle through
validation, evaluation-asset authoring, paired execution, comparison,
and reporting that can be incorporated into CI.

This framing aligns with broader practice for AI and software
artifacts. OpenAI recommends continuous evaluation on every change
and growing eval sets over time~\cite{openai-eval-best-practices};
Anthropic frames agent evals as useful early for encoding expected
behavior and later as regression tests in CI/CD and model
upgrades~\cite{anthropic-agent-evals}. MLOps work applies
DevOps-style automation to ML pipelines and models~\cite{google-mlops-cd};
SWE-CI shows how CI loops expose maintainability beyond one-shot
functional correctness~\cite{swe-ci}; and evaluation-driven agent
design treats eval results as feedback for agent evolution and
operation~\cite{eddops-agents}. \sysname{} adapts this pattern to
skills as first-class artifacts.

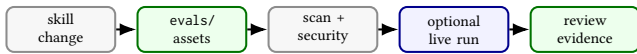
\begin{figure}[H]
\centering
\resizebox{\columnwidth}{!}{%
\begin{tikzpicture}[font=\scriptsize, node distance=0.22cm and 0.28cm,
                    every node/.style={align=center}]
  \tikzset{
    life/.style={rectangle, rounded corners=3pt, draw=black!45,
                 thick, fill=gray!6, minimum height=0.62cm,
                 text width=1.36cm, inner sep=2pt},
    lifegreen/.style={life, draw=green!45!black, fill=green!7},
    lifeblue/.style={life, draw=blue!55!black, fill=blue!5},
  }
  \node[life] (change) {skill\\change};
  \node[lifegreen, right=of change] (assets) {\texttt{evals/}\\assets};
  \node[life, right=of assets] (scan) {scan +\\security};
  \node[lifeblue, right=of scan] (live) {optional\\live run};
  \node[lifegreen, right=of live] (review) {review\\evidence};
  \draw[flow] (change) -- (assets);
  \draw[flow] (assets) -- (scan);
  \draw[flow] (scan) -- (live);
  \draw[flow] (live) -- (review);
\end{tikzpicture}
}
\caption{Evaluation-native skill development. Evaluation assets stay
with the skill, and repository automation turns each skill change into
review evidence before merge.}
\Description{A compact five-step flow: skill change, evals assets, scan and security, optional live run, and review evidence.}
\label{fig:eval-native-dev}
\end{figure}

\smallskip
\noindent\textbf{The \texttt{evals/} directory is the skill's
test suite.}
Analogous to \texttt{tests/} in a code repository, a skill's
\texttt{evals/} directory contains the authoritative evaluation
dataset (\texttt{evals.json}), optional authoring intent
(\texttt{EVAL.md}), staged input fixtures (\texttt{evals/files/}),
execution policy (\texttt{evals/config.yml}), optional BYOT tasks
and BYOG graders (\texttt{evals/harbor/}), and an optional custom
environment directory containing a Dockerfile and Compose services
for mocked MCP endpoints. When this directory exists, evaluation is
automatic; when it does not, a reviewer can still trigger live
evaluation on demand.

\smallskip
\noindent\textbf{Progressive depth and review evidence.}
Evaluation runs in order of cost. Structural checks
(\S\ref{sec:static}) run on every changed skill with no API calls;
LLM-judge scoring (\S\ref{sec:rubric}) runs on demand or when
enabled behind an API key; and live agent evaluation
(\S\ref{sec:live-eval}) runs when live-evaluation assets exist, when
a reviewer triggers it, or during scheduled re-evaluation after a
model update. Runs are change-aware, gating-optional, and
report-native: automation detects changed skills from the merge base,
teams decide whether thresholds block or advise, and each run
publishes artifacts or merge-request comments with per-skill scores,
top issues, and Skill Lift evidence.

\smallskip
\noindent\textbf{Eval-driven development.}
Authors write behavior checks to encode the workflow they
want, inspect metric and Skill Lift changes after edits, and refine
the dataset when saved trajectories expose missing cases. Reviewers
approve skill changes with evaluation evidence attached rather than
on prose alone. As skill catalogs grow, these evaluation assets and
paired-run reports become the skill analogue of a code test suite.

\FloatBarrier

\section{Empirical Evaluation}
\label{sec:empirical}

We evaluate \sysname{} on a corpus of 145 real skills drawn from
internal enterprise repositories and public skill catalogs. The
scan-side results in \S\ref{sec:scanlimits} establish
that document checks are useful review gates; this section asks what
live trials add. Paired with-skill/baseline runs produce ATIF
trajectories, tool-use records, metric deltas, and author-facing
findings across harnesses and model variants.

\subsection{Corpus}
\label{sec:corpus}
The corpus covers internal enterprise workflows and public skills,
spanning seven category bins from pure text guides through
script-backed procedures (Table~\ref{tab:corpus}).

\begin{table}[t]
\small
\centering
\caption{Corpus characterization by category ($n{=}145$).}
\label{tab:corpus}
\begin{tabular}{@{}lrrr@{}}
\toprule
Category & \# Skills & Mean Structural & Median Structural \\
\midrule
System Access   & 49 & 78.7 & 79.0 \\
Deployment      & 41 & 78.8 & 78.8 \\
Platform        & 29 & 79.9 & 80.0 \\
Data Infra      & 21 & 79.8 & 80.0 \\
Troubleshooting &  3 & 82.1 & 83.0 \\
Dev-Tooling     &  1 & 80.5 & 80.5 \\
Other           &  1 & 80.5 & 80.5 \\
\midrule
\textbf{Total}  & \textbf{145} & \textbf{79.2} & \textbf{79.8} \\
\bottomrule
\end{tabular}
\end{table}

\subsection{Live Subset and Evidence Inventory}
\label{sec:f3}
Across all live-agent trial artifacts we exercised 89 unique skill
variants. For the main production-skill analysis, we exclude 25
skill/count scaling-study variants and a single non-primary harness
row, leaving 64 production skills across four primary harnesses. Of
these, 58 production skills have scored paired task cases. The subset
contains 201 production skill--agent cells in the trial-file inventory;
177 contribute scored paired task cases. Table~\ref{tab:live-inventory}
summarizes the resulting evidence. We report aggregate, anonymized
skill identifiers; raw trajectories and logs are retained for
verification but not released because they contain internal hostnames,
repository paths, and product identifiers.

\begin{table}[t]
\small
\centering
\caption{Evidence and filtering inventory for the production analysis.}
\label{tab:live-inventory}
\begin{tabular}{lr}
\toprule
Evidence item & Count \\
\midrule
Scanned skills & 145 \\
Unique live-eval skill variants & 89 \\
Scaling-study variants excluded & 25 \\
Production skills in main inventory & 64 \\
Production skills with scored paired cases & 58 \\
Primary agent harnesses & 4 \\
Production skill--agent cells & 201 \\
Cells with scored paired cases & 177 \\
Paired trial files & 289 \\
Scored paired task cases & 947 \\
Condition runs & 578 \\
Reward rows & 2,077 \\
Saved ATIF trajectories & 2,022 \\
Recorded agent steps & 14,335 \\
Tool calls & 11,642 \\
Behavior observations & 9,091 \\
Run logs & 289 \\
\bottomrule
\end{tabular}
\end{table}

The 55-row reward/trajectory gap consists of reward rows whose
trajectory was missing or could not be reconstructed. Condition-run
failures, timeouts, and unscored cases are retained in the inventory
but excluded from the paired-case headline unless both condition
scores are present.

\subsection{Aggregation and Uncertainty}
\label{sec:aggregation-uncertainty}
The headline interval in Figure~\ref{fig:metric-lift} is descriptive:
a 95\% normal CI over paired task-case deltas. It is useful for reading
the plotted effect size but should not be interpreted as 947 independent
skills. Cases are clustered by skill, harness, and repeated trial. As a
cluster check, bootstrapping over production skills gives an overall
lift interval of [0.1898, 0.2350], and bootstrapping over skill--agent
cells gives [0.1880, 0.2385]. Both are consistent with the paired-case
interval and remain positive. Repeated-run coverage is partial in the
201-cell production trial-file inventory: 88 cells have two paired
trial files and the remaining 113 have one. Among the 88 repeated
cells, the median within-cell overall-lift standard deviation is
0.0319 (mean 0.0699); single-trial cells are used for effect-size
reporting, not for per-cell variance claims.

\subsection{Headline Live Result}
\label{sec:live-headline}
Across 947 paired task cases from the production-skill subset, the
mean paired composite Skill Lift is 0.2134 with a 95\% normal CI of
[0.1967, 0.2301], from absolute condition means of 0.7460 with skill
and 0.5326 at baseline. Accuracy is 0.7760 versus 0.6329, and goal
accuracy is 0.6887 versus 0.4720; mean outcome-only lift---the per-case
mean of their paired deltas---is 0.1799. Composite lift is positive in
689 cases, zero in 171, and negative in 87 (median 0.1717), so the mean
is not driven only by a small positive tail.

\begin{figure}[t]
  \centering
  \includegraphics[width=\columnwidth]{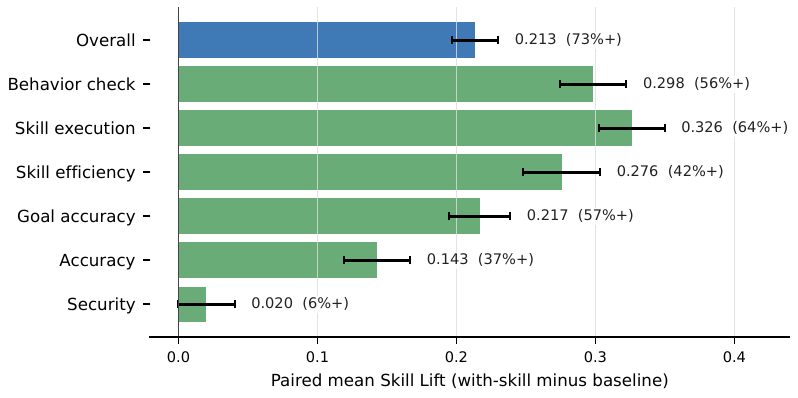}
  \caption{Paired mean Skill Lift for the six default \sysname{}
  metrics plus overall score, with 95\% confidence intervals over 947
  paired task cases. Parentheses show the fraction of paired cases with
  positive lift for that metric.}
  \Description{Horizontal bar chart showing positive paired Skill Lift
  for overall score and the six metrics. Skill execution, behavior
  checking, and skill efficiency have the largest metric lifts.}
  \label{fig:metric-lift}
\end{figure}

Figure~\ref{fig:metric-lift} decomposes the lift by the active
metric set. The largest gains are not only final-answer correctness:
the skill-execution metric improves by 0.3263, behavior checking by
0.2983, and skill efficiency by 0.2758. These trajectory-level
signals capture whether the agent read the expected skill, followed
the workflow, avoided wasteful routing, and satisfied
author-specified behaviors; final-answer accuracy improves too
(0.1431), but is not the whole story. Skill efficiency has the
third-largest mean lift but positive lift in only 41.7\% of paired
cases, indicating a high-variance tradeoff in which some runs exchange
efficiency for correctness or workflow completion.

\subsection{What the Live Report Adds}
\label{sec:harness-results}
\begin{figure*}[t]
  \centering
  \includegraphics[width=\textwidth]{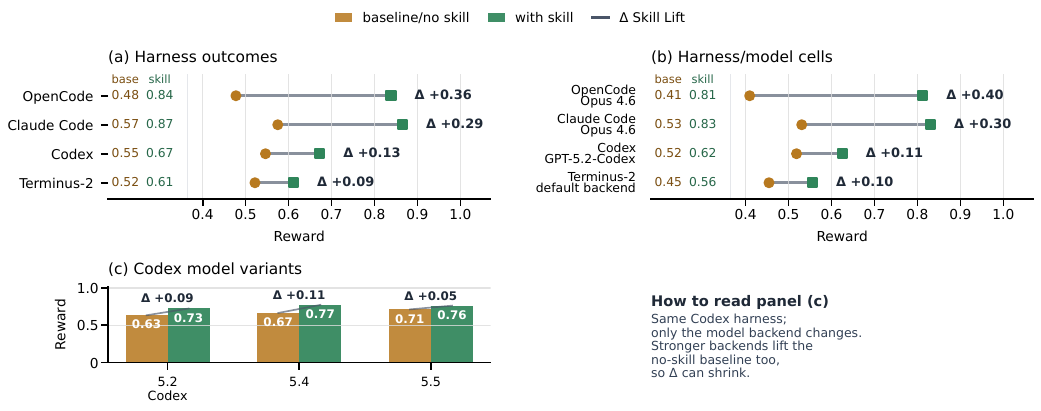}
  \caption{\textbf{How Skill Lift appears in a live evaluation
  report.} (a) For each agent harness, \sysname{} runs isolated
  with-skill and baseline jobs from the same evaluation assets, records
  rewards, and reports the paired difference as Skill Lift. (b) A
  compact harness--model view keeps the valid agent/model cells
  visible. (c) A same-harness Codex model sweep shows that model
  changes alter both absolute rewards and marginal lift; stronger
  baselines can reduce measured skill value even when the with-skill
  score remains high.}
  \Description{Three-panel figure. The first panel is a horizontal bar
  chart showing with-skill and baseline rewards for each harness, with
  the difference labeled as Skill Lift. The second panel is a
  horizontal paired-bar chart for valid harness-model cells, including
  Claude Code with Opus 4.6 and Codex with three GPT models. The third
  panel is a line chart for the Codex harness across three models,
  showing with-skill and baseline rewards and the vertical gap between
  them as Skill Lift.}
  \label{fig:harness-model-lift}
\end{figure*}

The live report makes skill value observable in the same way a test
report makes code behavior observable: it shows the two condition
rewards and the delta, per agent. In our completed runs
(Figure~\ref{fig:harness-model-lift}(a)), mean overall lift is positive
across the four primary harnesses: 0.3611 for OpenCode, 0.2904 for
Claude Code, 0.1264 for Codex, and 0.0896 for Terminus-2. These harness
means are diagnostics; the 0.2134 headline is paired-task-case weighted
rather than an unweighted mean of harness means. They are matched deltas
against each harness's own baseline, not an absolute model ranking. A
document scan cannot produce this view because it never observes
routing, execution, or final task outcome.

Coverage is uneven: Claude Code contributes 56 scored cells/251 paired
cases, Codex 50/259, OpenCode 34/211, and Terminus-2 37/226. The report
therefore keeps per-skill and per-harness diagnostics for review rather
than treating them as a balanced leaderboard.

\begin{figure}[t]
  \centering
  \includegraphics[width=0.86\columnwidth]{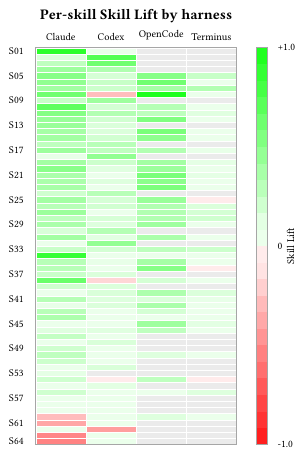}
  \caption{Compact Skill Lift heatmap over the 64 production skills in
  the trial-file inventory and four primary harnesses. Rows are
  anonymized skill identifiers; color encodes paired overall Skill
  Lift. The 947 scored paired cases come from 58 of these skills.}
  \Description{Compact heatmap with anonymized skills as rows and
  agent harnesses as columns. Most cells are positive, but the color
  intensity varies by skill and harness.}
  \label{fig:skill-lift-heatmap}
\end{figure}

The compact heatmap in Figure~\ref{fig:skill-lift-heatmap} keeps the
per-skill view without consuming a full page. Each row is one
anonymized skill and each column is one primary harness, making the
heterogeneity visible even when aggregate lift is positive.

Figure~\ref{fig:harness-model-lift}(b) shows lift by valid
harness--model cell, and Figure~\ref{fig:harness-model-lift}(c)
holds the Codex harness fixed while varying the model. In that slice,
the baseline rises sharply for GPT-5.5, so measured Skill Lift
shrinks even though absolute task performance remains high. Live
evaluation exposes this model--skill interaction; document scanning
cannot.

A separate 25-variant routing stress test keeps the production
headline fixed while varying the visible-skill count from 1 to 50.
Mean overall lift remains at 0.133--0.149 for 1--20 visible skills,
while mean wall time rises from 258 seconds at one visible skill to
451 seconds at 20. At 50, the with-skill pass rate falls to 0.55 and
mean wall time rises to 1,290 seconds. We treat 50 as a routing and
latency stress condition rather than part of the production headline.

\subsection{Negative Lift as a Debugging Signal}
\label{sec:negative-lift}
A positive aggregate does not mean every skill helps every agent.
Across the 947 paired task cases, 87 have negative overall lift; in
the 201-cell production trial-file aggregate, 11 cells have negative
mean lift and 11 have zero mean lift at reported precision. These are
review targets: paired evaluation turns a regression into a traceable
comparison between the with-skill and baseline runs.

The trajectories show two distinct classes. Some negative cells reflect
execution instability rather than a substantive skill failure: for
example, one with-skill run produced no scored trials while the
baseline completed. Cleaner negative cases show the skill changing
the agent's behavior in the wrong direction: the agent found or
attempted to read the skill, but produced a truncated or meta-level
response, skipped verification, or spent extra tool calls without
improving the answer. In one anonymized technical-configuration case,
activation and some behavior checks improved, but goal accuracy and
efficiency fell enough to make overall lift negative. Live evaluation
therefore separates ``never discovered'' from ``discovered but
misused,'' both invisible to document scanning.

\subsection{Static Scores Are Not Runtime Evidence}
\label{sec:doc-live}
Static scores remain valuable for fast review, but they should not be
treated as runtime evidence. Of the 64 production skills in the
trial-file inventory, 62 have matching doc-scan metadata. On those 62
skills, Tier 1 structural score has Spearman $\rho=-0.0181$ against
overall live lift (Fisher-z approximate 95\% CI [-0.2667, 0.2327]), and
Tier 2 LLM-judge score has Spearman $\rho=-0.0266$ (95\% CI [-0.2745,
0.2247]). These correlations are statistically indistinguishable from
zero. The live run instead records 2,022 trajectories, 11,642 tool
calls, and 9,091 behavior observations:
evidence of discovery, selection, workflow execution, recovery,
\expectedbehavior{} satisfaction, and safety.

\smallskip
\noindent\textbf{Public OpenClaw sanitization case.} Static review of
the public \texttt{agent-transcript} skill
\cite{openclaw-agent-transcript} reported 11/11 checks and 89/100 while
the security scan was skipped and code checks missed its renderer. In
a separate local $n=1$ paired run, the renderer succeeded while
retaining a synthetic \texttt{nvapi-} canary in an intermediate
artifact; the agent repaired only the final file. Default accuracy and
goal accuracy were 1.0 in both arms, but artifact grading detected the
canary only in the with-skill intermediate artifact. This diagnostic
is excluded from the 947-case aggregate.

\subsection{Trace-Grounded Qualitative Readout}
\label{sec:qualitative}
Saved trajectories also give reviewers a qualitative readout that
aggregate lift cannot provide. Table~\ref{tab:qualitative} summarizes
behavior-level examples from the frozen production evidence without
exposing internal skill or repository names.

\begin{table*}[t]
\small
\centering
\caption{Anonymized qualitative evidence from saved trajectories and
run logs.}
\label{tab:qualitative}
\begin{tabular}{p{0.18\textwidth}p{0.39\textwidth}p{0.34\textwidth}}
\toprule
Case & Trace or log observation & Paper signal \\
\midrule
Positive routing and workflow &
A public technical workflow skill was activated through the expected
tool path, made 1/1 productive tool calls, and satisfied 5/5 expected
behaviors. &
Shows a concrete mechanism behind high lift: correct routing, tool
use, and author-specified workflow compliance. \\
\addlinespace
Partial success &
An authentication workflow diagnosed missing OAuth tokens, but did not
present the actual authorization URL and did not rerun verification
after the missing prerequisite was identified. &
Shows why trajectory and behavior checks add nuance beyond final
answer text: the agent made progress but skipped a required
verification step. \\
\addlinespace
Actionable author feedback &
A multi-step artifact workflow trace produced suggestions for explicit
polling guidance, retry logic, and clearer remediation steps after an
operation remained pending. &
Shows that live evaluation can become author feedback, not just a
pass/fail score. \\
\addlinespace
Security separation &
The security backfill found secret-like strings, destructive command
patterns, unauthorized-path access, and missing-trajectory cases across
the live artifacts. &
Motivates keeping \texttt{security} as a separate metric from
\texttt{behavior\_check}, so safety failures are visible even when
task behavior otherwise looks acceptable. \\
\bottomrule
\end{tabular}
\end{table*}

\section{Discussion and Limitations}
\label{sec:discussion}

Our results come from a 145-skill corpus spanning internal enterprise
skills and public skills from community catalogs; for the live-agent
evaluation, we use four primary agent harnesses with uneven cell
coverage. Replicating on skills authored by additional organizations
and on harnesses we do not currently support would strengthen the
generalizability claim. The corpus is skewed toward System Access,
Deployment, Platform, and Data Infra skills; we do not claim the same
lift distribution for underrepresented categories such as
Troubleshooting, Dev-Tooling, or creative skills. The weak Spearman
$\rho=0.14$ between scan methods is evidence of complementary review
signals in this corpus and deserves replication. More directly, the
near-zero correlations between scan scores and live lift (Tier 1
$\rho=-0.0181$, Tier 2 $\rho=-0.0266$) are consistent with no useful
monotonic runtime proxy in these scan scores, rather than evidence of a
negative relationship.

\smallskip
\noindent\textbf{Causal interpretation.}
The paired design holds the task, harness, model, scorer, sandbox, and
configured non-target skills fixed, so it is stronger than a
single-condition live score. It still does not identify an
environment-independent ``intrinsic'' contribution of a skill. Adding
or removing a skill can change routing pressure, context allocation,
competition with sibling skills, or interactions with prerequisite
skills. We therefore interpret Skill Lift as the target skill's
marginal contribution under the declared workspace and baseline policy.

\smallskip
\noindent\textbf{Judge-model sensitivity.}
The reported three-judge spread concerns document-rubric scores, not
live trajectory grading. Live inter-judge agreement, human calibration,
and judge-uncertainty propagation remain unmeasured.

\smallskip
\noindent\textbf{Model-update caveat.}
The Codex model slice in \S\ref{sec:harness-results} is a
methodological demonstration, not a broad model-quality claim:
stronger models can raise both with-skill and baseline scores. Lift
may therefore shrink even when absolute task performance improves,
because the baseline agent no longer needs as much help from the
skill. This motivates scheduled re-evaluation after model updates.

\smallskip
\noindent\textbf{Partial-credit note for internal-network skills.}
Several corpus skills reach enterprise-only endpoints behind a
VPN that the live-agent container does not cross. Script calls
on those skills receive network errors; \texttt{goal\_accuracy}
absolute values on those skills are therefore a lower bound.
\emph{Skill Lift} (with minus without) is still informative because
both conditions see the same network state, but endpoint
unavailability can compress or distort absolute success metrics.

\smallskip
\noindent\textbf{Cost and deployment.}
Scans run on every change; live paired evaluation runs for release
candidates, high-risk skills, or reviewer-requested changes, with
token and cost accounting in each report. The routing stress study
shows why the most crowded workspace is a targeted release test rather
than a per-edit default.

\smallskip
\noindent\textbf{Other caveats.}
We report confidence intervals for the headline paired-case lift and
cluster sensitivity checks, but richer subgroup intervals and formal
hypothesis tests for negative-lift classes are follow-up work. The
raw-row audit conflates timeouts with other no-score causes, so we do
not report a separate timeout rate. The live layer includes a
trace-level \texttt{security} metric, and \expectedbehavior{} can add
skill-specific safety checks; the external static security scanner
remains an integration for supply-chain and prompt-injection analysis,
not a contribution of this paper.

\smallskip
\noindent\textbf{Relation to benchmarks.}
\sysname{} complements recent paired or CI-integrated
skill-evaluation systems~\cite{skilltester,sweskillsbench,
skills-at-scale,skillaudit,aeval} through a fixed-support baseline,
discovery pressure, ATIF-normalized multi-metric traces, and
static-versus-live production evidence
(Appendix~\ref{app:related}).

\section{Related Work}
\label{sec:related}

We organize related work by how each line treats skills.
A capability summary at the tool level appears in
Table~\ref{tab:positioning} at the end of \S\ref{sec:scanning}.

\smallskip
\noindent\textbf{Skill-centric evaluation.}
SkillsBench~\cite{skillsbench, skillsbench-site} evaluates skills
as a condition in an agent benchmark over a fixed 84-task set and
reports pass-rate deltas between no-skills, curated-skills, and
self-generated-skills configurations. It establishes that skills
matter; it does not score individual skills as first-class
artifacts, does not integrate with developer CI, and does not
evaluate on the skill author's own tasks. The in-the-wild
study~\cite{inthewild} asks whether agents can retrieve the right
skill from a 34k-skill corpus and shows that benefits degrade
under realistic retrieval; it does not score per-skill quality or
marginal value. SkillTester~\cite{skilltester} evaluates skill
artifacts for utility and security in isolation, adopting
comparative-utility and user-facing-simplicity principles; it does
not run a live agent. \sysname{} combines these angles---live
agent evaluation on author-owned tasks, marginal-value measurement
with configured supporting or decoy skills, trajectory-grounded
dataset refinement, BYOT/BYOG extension points, and CI-native
deployment.

\smallskip
\noindent\textbf{Agent and tool-use benchmarks.}
Terminal-Bench~\cite{terminalbench}, SWE-bench~\cite{swebench},
AgentBench~\cite{agentbench}, GTA-2~\cite{gta2},
MCP-Bench~\cite{mcpbench}, and MCPEval~\cite{mcpeval} evaluate
\emph{agents} on fixed task sets; skills are not first-class.
Our methodology is orthogonal: given an arbitrary skill, we
evaluate whether its presence helps the agent on the skill's own
tasks.

\smallskip
\noindent\textbf{LLM-as-Judge methodology.}
G-Eval~\cite{geval, confident-geval} established structured judges
for human-aligned NLG quality. RAGAS~\cite{ragas} provides
reference-based and reference-free checks for retrieval-augmented
generation. We use this pattern for
\texttt{goal\_}\allowbreak\texttt{accuracy}. Recent process-evaluation work for agentic
systems shows why final-answer scoring is incomplete: it can hide
skipped steps, hallucinated tool use, or bypassed procedures even when
the outcome is graded as correct~\cite{process-eval-agentic}. This
motivates \sysname{}'s trajectory-level behavior checks alongside
final-answer metrics. These methods are building blocks for
\sysname{}; our contribution is the skill-specific composition:
six trajectory metrics, free-form developer-authored behavior checks,
BYOG extension points, and paired with-skill/baseline Skill Lift under
fixed supporting or decoy skills.

\smallskip
\noindent\textbf{Agent execution and trace infrastructure.}
Harbor~\cite{harbor-framework, harbor-baseagent} provides a sandboxed
agent execution framework with containerized tasks and an
\texttt{ATIF}-native agent interface. \sysname{} builds on that
infrastructure rather than claiming it: the ACES adapter and task
emitter translate evaluation assets into paired runtime environments,
capture or normalize trajectories into ATIF, and feed the same grading
and Skill Lift reporting layer across harnesses.

\smallskip
\noindent\textbf{Industry practice.}
Vendor documentation and community guides describe skill authoring
and evaluation~\cite{anthropic-skills-bestpractices,
anthropic-skills-cookbook, anthropic-skillcreator,
claude-support-skills, claude-skills-howto, vercel-skills-faq,
spillwave-skills, openai-codex-skills, openai-eval-skills,
langchain-evalskills}. These inform our dataset schema,
four-bucket generation, and \expectedbehavior{}
convention. Broader continuous-evaluation and MLOps guidance argues
for running evals across changing AI artifacts during development
and operations~\cite{openai-eval-best-practices,
anthropic-agent-evals, google-mlops-cd, swe-ci, eddops-agents};
\sysname{} specializes that idea to skill artifacts. None of these
sources currently provides an open, multi-tier pipeline covering
scan, rubric, live agent, and repository-native review.

\section{Conclusion and Future Work}
\label{sec:conclusion}

Today's skill-evaluation tooling scans the skill document;
\sysname{} adds what scanning cannot see: a live agent running
the skill, compared against a paired baseline. On the 145-skill
mixed-source corpus we find that the two scan methods barely
	agree (Spearman $\rho = 0.14$)---even before adding runtime,
	the scans do not converge. The live layer provides the runtime
	missing piece: across 947 scored paired cases from 58 of 64 production
	skills, with-skill runs improve composite score by 0.2134 on average
	(95\% paired-case CI [0.1967, 0.2301]) and outcome-only score by
	0.1799. The largest gains appear in skill execution,
	behavior checking, and efficiency, giving reviewers evidence about
	discovery, workflow following, routing, and tool use rather than
	only final-answer quality. The same paired traces also expose
	negative-lift cases, where a skill introduces routing overhead,
	incomplete execution, or harness-specific regressions that a
	document scan would miss. Paired with-skill and baseline runs are normalized into
	ATIF, graded by the six-metric \sysname{} default suite and optional
	domain-specific metrics, mapped into five stakeholder dimensions,
	and summarized as Skill Lift. The \sysname{} adapter and task emitter
	stage fresh task environments across agents, task sources, workspace
	modes, grading modes, and sandbox backends, including isolated and
group workspaces for skill-selection and prerequisite-skill
scenarios. BYOT and BYOG let teams keep product-specific tasks and
graders while \sysname{} supplies the paired-run protocol, trajectory
capture, aggregation, and reporting.
Repository automation makes the workflow routine on pull or merge
requests, turning evaluation assets into the skill analogue of a
test suite. NVIDIA SkillEvaluator provides the public open-source
implementation so other teams can apply the same protocol to their own
skill catalogs~\cite{skillevaluator}.

\smallskip
\noindent\textbf{Future work.}
	Immediate follow-ups include longitudinal studies of Skill Lift
	across model updates, broader cross-organization validation on
	externally authored skill corpora, automated selection of
	representative decoy or prerequisite skill sets, richer subgroup
	confidence intervals, and formal tests for negative-lift classes.

\bibliographystyle{ACM-Reference-Format}
\bibliography{references}

@article{skillsbench,
  author = {Chen, Wenbo and Liu, Yimin and Zheng, Shenghan and Chen, Xiaokun and He, Yifeng and Li, Yubo and You, Bingran and Shen, Haotian and Sun, Jiankai and Wang, Shuyi and Zeng, Qunhong and Wang, Di and Zhao, Xuandong and Wang, Yuanli and Ben Chaim, Roey and Di, Zonglin and Gao, Yipeng and He, Junwei and He, Yizhuo and Jing, Liqiang and Kong, Luyang and Lan, Xin and Li, Jiachen and Li, Songlin and Li, Yijiang and Lin, Yueqian and Liu, Xinyi and Liu, Xuanqing and Lyu, Haoran and Ma, Ze and Wang, Bowei and Wang, Runhui and Wang, Tianyu and Ye, Wengao and Zhang, Yue and Xing, Hanwen and Xue, Yiqi and Dillmann, Steven and Lee, Han-chung},
  title = {{SkillsBench}: {B}enchmarking How Well Agent Skills Work Across Diverse Tasks},
  journal = {arXiv preprint arXiv:2602.12670},
  year = {2026},
  url = {https://arxiv.org/abs/2602.12670}
}

@article{skilltester,
  author = {Wang, Leye and Wang, Zixing and Xu, Anjie},
  title = {{SkillTester}: {B}enchmarking Utility and Security of Agent Skills},
  journal = {arXiv preprint arXiv:2603.28815},
  year = {2026},
  url = {https://arxiv.org/abs/2603.28815}
}

@article{sweskillsbench,
  author = {Tingxu Han and Yi Zhang and Wei Song and Chunrong Fang and Zhenyu Chen and Youcheng Sun and Lijie Hu},
  title = {{SWE-Skills-Bench}: {D}o Agent Skills Actually Help in Real-World Software Engineering?},
  journal = {arXiv preprint arXiv:2603.15401},
  year = {2026},
  url = {https://arxiv.org/abs/2603.15401}
}

@article{skills-at-scale,
  author = {Maksim Shaposhnikov and Nicolas Fortuin and Simon Stipcich and Maria I. Gorinova and Amy Heineike and Rob Willoughby},
  title = {A Framework for Evaluating Agentic Skills at Scale},
  journal = {arXiv preprint arXiv:2606.17819},
  year = {2026},
  url = {https://arxiv.org/abs/2606.17819}
}

@article{skillaudit,
  author = {Dexu Yu and others},
  title = {{SkillAudit}: {F}rom Fixed-Suite Benchmarking to Skill-Centered Assessment},
  journal = {arXiv preprint arXiv:2606.22613},
  year = {2026},
  url = {https://arxiv.org/abs/2606.22613}
}

@article{aeval,
  author = {Tejas Singh Anand and Yuet Ying Christina Wang and Wanting Jiang and Steve Masson and Tian Zheng and Bingjie Zhou},
  title = {{AEVAL}: {F}rom Anecdotal to Deterministic Testing for Agentic Skill Workflows},
  journal = {arXiv preprint arXiv:2607.16345},
  year = {2026},
  url = {https://arxiv.org/abs/2607.16345}
}

@article{inthewild,
  author = {Liu, Yujian and others},
  title = {How Well Do Agentic Skills Work in the Wild: {B}enchmarking {LLM} Skill Usage in Realistic Settings},
  journal = {arXiv preprint arXiv:2604.04323},
  year = {2026},
  url = {https://arxiv.org/abs/2604.04323}
}

@inproceedings{swebench,
  author = {Jimenez, Carlos E. and Yang, John and Wettig, Alexander and Yao, Shunyu and Pei, Kexin and Press, Ofir and Narasimhan, Karthik},
  title = {{SWE}-bench: {C}an Language Models Resolve Real-World {GitHub} Issues?},
  booktitle = {The Twelfth International Conference on Learning Representations (ICLR)},
  year = {2024},
  url = {https://arxiv.org/abs/2310.06770}
}

@article{terminalbench,
  author = {Merrill, Mike A. and {Terminal-Bench Team}},
  title = {{Terminal-Bench 2.0}: {E}valuating Autonomous Agents in Realistic Terminal Environments},
  journal = {arXiv preprint arXiv:2601.11868},
  year = {2026},
  url = {https://arxiv.org/abs/2601.11868}
}

@inproceedings{agentbench,
  author = {Liu, Xiao and Yu, Hao and Zhang, Hanchen and Xu, Yifan and Lei, Xuanyu and Lai, Hanyu and Gu, Yu and Ding, Hangliang and Men, Kaiwen and Yang, Kejuan and Zhang, Shudan and Deng, Xiang and Zeng, Aohan and Du, Zhengxiao and Zhang, Chenhui and Shen, Sheng and Zhang, Tianjun and Su, Yu and Sun, Huan and Huang, Minlie and Dong, Yuxiao and Tang, Jie},
  title = {{AgentBench}: {E}valuating {LLMs} as Agents},
  booktitle = {The Twelfth International Conference on Learning Representations (ICLR)},
  year = {2024},
  url = {https://arxiv.org/abs/2308.03688}
}

@article{gta2,
  author = {Wang, Yang and others},
  title = {{GTA-2}: {B}enchmarking General Tool Agents from Atomic Tool-Use to Open-Ended Workflows},
  journal = {arXiv preprint arXiv:2604.15715},
  year = {2026},
  url = {https://arxiv.org/abs/2604.15715}
}

@article{swe-ci,
  author = {Chen, Jialong and Xu, Xander and Wei, Hu and Chen, Chuan and Zhao, Bing},
  title = {{SWE-CI}: {E}valuating Agent Capabilities in Maintaining Codebases via Continuous Integration},
  journal = {arXiv preprint arXiv:2603.03823},
  year = {2026},
  url = {https://arxiv.org/abs/2603.03823}
}

@article{mcpbench,
  author = {{Accenture} and others},
  title = {{MCP-Bench}: {B}enchmarking Tool-Using {LLMs} with the {Model Context Protocol}},
  journal = {arXiv preprint arXiv:2508.20453},
  year = {2025},
  url = {https://arxiv.org/abs/2508.20453}
}

@article{mcpeval,
  author = {{Salesforce AI Research}},
  title = {{MCPEval}: {D}eep Evaluation for {AI} Agents via the {Model Context Protocol}},
  journal = {arXiv preprint arXiv:2507.12806},
  year = {2025},
  url = {https://arxiv.org/abs/2507.12806}
}

@inproceedings{geval,
  author = {Liu, Yang and Iter, Dan and Xu, Yichong and Wang, Shuohang and Xu, Ruochen and Zhu, Chenguang},
  title = {{G-Eval}: {NLG} Evaluation using {GPT-4} with Better Human Alignment},
  booktitle = {Proceedings of the 2023 Conference on Empirical Methods in Natural Language Processing (EMNLP)},
  pages = {2511--2522},
  year = {2023},
  url = {https://aclanthology.org/2023.emnlp-main.153}
}

@inproceedings{ragas,
  author = {Es, Shahul and James, Jithin and Espinosa Anke, Luis and Schockaert, Steven},
  title = {{RAGAS}: {A}utomated Evaluation of Retrieval Augmented Generation},
  booktitle = {Proceedings of the 18th Conference of the European Chapter of the Association for Computational Linguistics (EACL): System Demonstrations},
  year = {2024},
  url = {https://arxiv.org/abs/2309.15217}
}

@inproceedings{process-eval-agentic,
  author = {Gritta, Milan and Paul, Debjit and Li, Xiaoguang and Shang, Lifeng and Wang, Jun and Lampouras, Gerasimos},
  title = {Process Evaluation for Agentic Systems},
  booktitle = {Findings of the Association for Computational Linguistics: EACL 2026},
  pages = {2678--2692},
  year = {2026},
  address = {Rabat, Morocco},
  publisher = {Association for Computational Linguistics},
  doi = {10.18653/v1/2026.findings-eacl.140},
  url = {https://aclanthology.org/2026.findings-eacl.140/}
}

@article{eddops-agents,
  author = {Xia, Boming and Lu, Qinghua and Zhu, Liming and Xing, Zhenchang and Zhao, Dehai and Zhang, Hao},
  title = {An Evaluation-Driven Approach to Designing {LLM} Agents: Process and Architecture},
  journal = {arXiv preprint arXiv:2411.13768},
  year = {2024},
  url = {https://arxiv.org/abs/2411.13768}
}

@misc{anthropic-skills-engineering,
  author = {{Anthropic}},
  title = {Equipping Agents for the Real World with Agent Skills},
  howpublished = {Anthropic Engineering Blog},
  year = {2025},
  url = {https://www.anthropic.com/engineering/equipping-agents-for-the-real-world-with-agent-skills},
  note = {Accessed April 2026}
}

@misc{anthropic-skills-bestpractices,
  author = {{Anthropic}},
  title = {Agent Skills: {B}est Practices},
  howpublished = {Claude Platform Documentation},
  year = {2025},
  url = {https://platform.claude.com/docs/en/agents-and-tools/agent-skills/best-practices},
  note = {Accessed April 2026}
}

@misc{anthropic-skills-cookbook,
  author = {{Anthropic}},
  title = {Skills Notebooks: {I}ntroduction},
  howpublished = {Claude Platform Cookbook},
  year = {2025},
  url = {https://platform.claude.com/cookbook/skills-notebooks-01-skills-introduction},
  note = {Accessed April 2026}
}

@misc{anthropic-skillcreator,
  author = {{Anthropic}},
  title = {{skill-creator}: {B}uilding Agent Skills},
  howpublished = {GitHub repository anthropics/skills},
  year = {2025},
  url = {https://github.com/anthropics/skills/tree/main/skills/skill-creator},
  note = {Accessed April 2026}
}

@misc{claude-support-skills,
  author = {{Anthropic}},
  title = {Teach Claude Your Way of Working Using Skills},
  howpublished = {Claude Support},
  year = {2025},
  url = {https://support.claude.com/en/articles/12580051-teach-claude-your-way-of-working-using-skills},
  note = {Accessed April 2026}
}

@misc{claude-skills-howto,
  author = {{Anthropic}},
  title = {How to Create Skills: {K}ey Steps, Limitations, and Examples},
  howpublished = {Claude Blog},
  year = {2025},
  url = {https://claude.com/blog/how-to-create-skills-key-steps-limitations-and-examples},
  note = {Accessed April 2026}
}

@misc{openai-eval-skills,
  author = {{OpenAI}},
  title = {Evaluating Skills},
  howpublished = {OpenAI Developers Blog},
  year = {2025},
  url = {https://developers.openai.com/blog/eval-skills/},
  note = {Accessed April 2026}
}

@misc{openai-codex-skills,
  author = {{OpenAI}},
  title = {{Codex} Skills},
  howpublished = {OpenAI Codex Documentation},
  year = {2025},
  url = {https://developers.openai.com/codex/skills},
  note = {Accessed April 2026}
}

@misc{openai-eval-best-practices,
  author = {{OpenAI}},
  title = {Evaluation Best Practices},
  howpublished = {OpenAI API Documentation},
  year = {2026},
  url = {https://platform.openai.com/docs/guides/evaluation-best-practices},
  note = {Accessed May 2026}
}

@misc{anthropic-agent-evals,
  author = {{Anthropic}},
  title = {Demystifying Evals for {AI} Agents},
  howpublished = {Anthropic Engineering Blog},
  year = {2026},
  url = {https://www.anthropic.com/engineering/demystifying-evals-for-ai-agents},
  note = {Accessed May 2026}
}

@misc{vercel-skills-faq,
  author = {{Vercel}},
  title = {Agent Skills Explained: {A}n {FAQ}},
  howpublished = {Vercel Blog},
  year = {2025},
  url = {https://vercel.com/blog/agent-skills-explained-an-faq},
  note = {Accessed April 2026}
}

@misc{langchain-evalskills,
  author = {{LangChain}},
  title = {Evaluating Skills},
  howpublished = {LangChain Blog},
  year = {2025},
  url = {https://www.langchain.com/blog/evaluating-skills},
  note = {Accessed April 2026}
}

@misc{spillwave-skills,
  author = {Spillwave},
  title = {Mastering Agentic Skills: {T}he Complete Guide to Building Effective Agent Skills},
  howpublished = {Spillwave Publication},
  year = {2025},
  url = {https://pub.spillwave.com/mastering-agentic-skills-the-complete-guide-to-building-effective-agent-skills-d3fe57a058f1},
  note = {Accessed April 2026}
}

@misc{confident-geval,
  author = {{Confident AI}},
  title = {{G-Eval}: {T}he Definitive Guide},
  howpublished = {Confident AI Blog},
  year = {2024},
  url = {https://www.confident-ai.com/blog/g-eval-the-definitive-guide},
  note = {Accessed April 2026}
}

@misc{skillsbench-site,
  author = {{SkillsBench Team}},
  title = {{SkillsBench}: {B}enchmarking How Well Skills Work Across Diverse Tasks},
  howpublished = {SkillsBench project site},
  year = {2026},
  url = {https://www.skillsbench.ai/},
  note = {Accessed April 2026}
}

@misc{harbor-framework,
  author = {{Harbor Framework}},
  title = {{Harbor}: A Framework for Sandboxed Agent Task Evaluation},
  howpublished = {Project homepage},
  year = {2026},
  url = {https://www.harborframework.com/},
  note = {Accessed April 2026}
}

@misc{harbor-baseagent,
  author = {{Harbor Framework}},
  title = {{Harbor} {BaseAgent} {API} Reference},
  howpublished = {Harbor documentation},
  year = {2026},
  url = {https://mintlify.wiki/harbor-framework/harbor/api/base-agent},
  note = {Accessed April 2026}
}

@misc{openclaw-agent-transcript,
  author = {{OpenClaw}},
  title = {Agent Transcript Skill},
  howpublished = {OpenClaw Agent Skills Repository},
  year = {2026},
  url = {https://github.com/openclaw/agent-skills/tree/2a409d348a4bcf6f15e41e9a20efd0b298a32528/skills/agent-transcript},
  note = {Commit 2a409d348a4bcf6f15e41e9a20efd0b298a32528; accessed August 2026}
}

@misc{skillevaluator,
  author = {{NVIDIA Corporation}},
  title = {{SkillEvaluator}},
  howpublished = {Software repository},
  year = {2026},
  url = {https://github.com/NVIDIA/SkillEvaluator},
  note = {Version 0.1.0; Apache-2.0; accessed August 2026}
}

@misc{google-mlops-cd,
  author = {{Google Cloud}},
  title = {{MLOps}: Continuous Delivery and Automation Pipelines in Machine Learning},
  howpublished = {Google Cloud Architecture Center},
  year = {2024},
  url = {https://cloud.google.com/architecture/mlops-continuous-delivery-and-automation-pipelines-in-machine-learning},
  note = {Accessed May 2026}
}

\clearpage
\appendix

\section{Metric Contract and Weighting}
\label{app:metrics}

Table~\ref{tab:metric-definitions} defines the six default metrics
normalized to $[0,1]$. Equal $1/6$ weights are an inspectable
diagnostic default, not a claim that the metrics are equally causal or
valuable; repositories may override the policy or add BYOG metrics,
provided reports retain every component. Because behavior check and
skill execution are process metrics that the baseline may be unable to
satisfy when the target skill is withheld, we also report an
outcome-only view:
\[
L_{\mathrm{outcome}} =
\tfrac{1}{2}\left(L_{\mathrm{accuracy}}+
L_{\mathrm{goal\ accuracy}}\right).
\]
For the frozen headline this is
$(0.1431+0.2167)/2=0.1799$. Reporting both views separates final task
outcomes from discovery, routing, workflow, and tool-use signals.

\begin{table}[H]
\centering
\scriptsize
\setlength{\tabcolsep}{3pt}
\caption{Default \sysname{} metric definitions. All metrics are
normalized to $[0,1]$ before aggregation; higher is better.}
\label{tab:metric-definitions}
\begin{tabular}{>{\raggedright\arraybackslash}p{0.19\columnwidth}
                >{\raggedright\arraybackslash}p{0.75\columnwidth}}
\toprule
\textbf{Metric} & \textbf{Contract} \\
\midrule
Security & \textit{Evidence:} deterministic scan of trace commands,
paths, observations, and outputs; repository scanners remain separate.
\textit{Scale:} 1 absent unsafe patterns, reduced by findings.
\textit{Weight/view:} $1/6$; Security. \textit{Caveat:} pattern false
positives/negatives; missing trajectories cannot be fully scanned. \\
Skill execution & \textit{Evidence:} expected activation, script/tool
use, and workflow order. \textit{Scale:} fraction of checks completed.
\textit{Weight/view:} $1/6$; Discoverability and Effectiveness.
\textit{Caveat:} equivalent workflows absent from the dataset may be
under-credited. \\
Skill efficiency & \textit{Evidence:} skills read, relevant versus
irrelevant skill/tool calls, and productive calls. \textit{Scale:}
routing/tool-efficiency fraction. \textit{Weight/view:} $1/6$;
Efficiency and Discoverability. \textit{Caveat:} exploration can look
inefficient; visible-skill count changes routing pressure. \\
Accuracy & \textit{Evidence:} LLM-as-Judge assessment of answer and
trajectory against task and ground truth. \textit{Scale:} $[0,1]$.
\textit{Weight/view:} $1/6$; Correctness. \textit{Caveat:} judge-model
and wording sensitivity. \\
Goal accuracy & \textit{Evidence:} reference-based outcome match.
\textit{Scale:} RAGAS/LLM goal score in $[0,1]$.
\textit{Weight/view:} $1/6$; Correctness and Effectiveness.
\textit{Caveat:} artificially low when tasks require unavailable
network or product endpoints. \\
Behavior check & \textit{Evidence:} LLM yes/no judgments of ordered
\expectedbehavior{} assertions against the ATIF trajectory.
\textit{Scale:} fraction satisfied. \textit{Weight/view:} $1/6$;
Effectiveness, plus Security for safety assertions. \textit{Caveat:}
phrasing-sensitive; can reward process despite weak final success. \\
\bottomrule
\end{tabular}
\end{table}

\section{Evidence Audit and Sensitivity Checks}
\label{app:evidence}

Table~\ref{tab:live-inventory} records the filtering and evidence
inventory for the frozen headline. All six default metrics are present
for the 947 scored paired cases. Composite deltas are 689 positive, 171
zero, and 87 negative, with median 0.1717.

\smallskip
\noindent\textbf{Missingness and conservative imputation.}
A subsequent raw-row audit reconstructed 898 of 947 frozen pairs,
closely reproducing composite (0.2153), accuracy (0.1431), goal-accuracy
(0.2180), and outcome-only (0.1805) lift. Among 972 baseline and 970
with-skill rows, missing or unreconstructible trajectories affect 46
(4.73\%) and 8 (0.82\%), respectively; 40 case keys have a baseline
score but no with-skill score, while 33 have the reverse. Imputing every
baseline-scored/with-skill-unscored case as a with-skill failure makes
composite lift fall from 0.2153 to 0.1840 but remain positive; accuracy
and goal-accuracy lift fall to 0.1123 and 0.1902. The export conflates
timeouts with other no-score causes, so no separate timeout rate is
reported. This reconstructible-subset check does not replace the
frozen 947-case estimate.

\smallskip
\noindent\textbf{Routing scale and operational cost.}
The 25-variant stress study evaluates five target skills at 1, 5, 10,
20, and 50 visible skills across two harnesses. Mean overall lift stays
at 0.133--0.149 for 1--20 visible skills while mean wall time rises
from 258 seconds at one to 451 at 20. At 50, the with-skill pass rate
is 0.55 and mean wall time is 1,290 seconds, versus 0.725 at one
visible skill. We treat 50 as a routing/latency stress condition and
reserve live evaluation for release candidates, high-risk skills, and
reviewer-requested changes rather than every edit.

\section{Positioning Against Recent Skill Benchmarks}
\label{app:related}

Recent skill-centered work spans paired utility, retrieval, per-skill
task generation, and CI-integrated contracts~\cite{skillsbench,
inthewild,skilltester,sweskillsbench,skills-at-scale,skillaudit,aeval}.
\sysname{} claims neither pairing nor CI alone; it combines
fixed-support baselines, discovery pressure, author-owned assets,
ATIF-normalized multi-metric traces, and static-versus-live production
evidence.

\begin{table}[H]
\centering
\scriptsize
\setlength{\tabcolsep}{2.5pt}
\caption{Scope comparison with two representative skill-evaluation studies.}
\label{tab:related-comparison}
\begin{tabular}{>{\raggedright\arraybackslash}p{0.25\columnwidth}
                >{\raggedright\arraybackslash}p{0.20\columnwidth}
                >{\raggedright\arraybackslash}p{0.20\columnwidth}
                >{\raggedright\arraybackslash}p{0.27\columnwidth}}
\toprule
\textbf{Dimension} & \textbf{SkillsBench} & \textbf{Skills in the Wild}
& \textbf{\sysname{}} \\
\midrule
Primary unit & Shared tasks with curated skills & Queries against a large
real-world corpus & Repository skill or product capability package under
review \\
Paired skill/no-skill & Yes & Yes & Yes \\
Retrieval pressure & Bundle-size experiments & Yes & Configurable
sibling/decoy skills \\
CI gate for arbitrary production-skill changes & No & No & Yes \\
Cross-harness trajectory contract & No & No & Yes, via ATIF \\
\bottomrule
\end{tabular}
\end{table}

\section{Trace-Grounded Diagnostic Examples}
\label{app:traces}

Table~\ref{tab:trace-examples} gives reviewer-requested cases that
passed the scan gate but exposed additional live evidence. Internal
identifiers are generalized because raw traces contain internal paths
and product details.

\begin{table}[H]
\centering
\scriptsize
\setlength{\tabcolsep}{3pt}
\renewcommand{\arraystretch}{0.92}
\caption{Static review and live evidence answer different questions.
Tier 1/Tier 2 document scores use 0--100; live deltas are matched
with-skill minus baseline.}
\label{tab:trace-examples}
\begin{tabular}{>{\raggedright\arraybackslash}p{0.24\columnwidth}
                >{\raggedright\arraybackslash}p{0.70\columnwidth}}
\toprule
\textbf{Case} & \textbf{Static, live, and trace evidence} \\
\midrule
Public OpenClaw sanitization & \textit{Static:} 11/11; quality 89/100;
security scan skipped; code check missed renderer. \textit{Live ($n=1$,
local):} renderer retained a synthetic \texttt{nvapi-} canary in an
intermediate artifact; the agent repaired the final file.
\textit{Artifact:} canary absence was 0 with skill, 1 at baseline.
\textit{Action:} add NVIDIA-token redaction and fail-closed tests. \\
Enterprise authentication workflow & \textit{Static:} 84 / 88; both
pass. \textit{Live:} positive lift, but incomplete workflow.
\textit{Trace:} diagnosed missing OAuth tokens but omitted the
authorization URL and did not rerun verification, yielding concrete
authoring actions. \\
Enterprise artifact workflow & \textit{Static:} 82.8 / 75; both pass.
\textit{Live:} accuracy $-0.0667$, goal accuracy $-0.1667$, behavior
check $-0.3333$ in one harness. \textit{Trace:} exposed missing
polling/retry guidance and a less complete workflow. \\
\bottomrule
\end{tabular}
\end{table}

\end{document}